\documentclass{article}

\PassOptionsToPackage{numbers, compress}{natbib}

\usepackage[preprint]{neurips_2021}

\usepackage[utf8]{inputenc} 
\usepackage[T1]{fontenc}    
\usepackage{hyperref}       
\usepackage{url}            
\usepackage{booktabs}       
\usepackage{amsfonts}       
\usepackage{nicefrac}       
\usepackage{microtype}      
\usepackage{xcolor}         
\usepackage{graphicx}
\usepackage{caption}
\usepackage[linesnumbered, ruled,vlined]{algorithm2e}
\usepackage{amsmath}
\usepackage{subfigure}
\usepackage{array}
\usepackage{wrapfig}
\usepackage{enumitem}
\usepackage{multirow}
\usepackage{lscape}
\usepackage[title]{appendix}

\title{FedCCEA : A Practical Approach of Client Contribution Evaluation for Federated Learning}
\author{Sung Kuk Shyn$^*$
    \\Department of Artificial Intelligence \\
    Sungkyunkwan University\\%
  Suwon, South Korea \\%
  \texttt{davidshyn@skku.edu} 
  \And  Donghee Kim \\
  College of Computing\\
  Sungkyunkwan University\\%
  Suwon, South Korea \\%
  \texttt{ym.dhkim@skku.edu}
  \And Kwangsu Kim \\
  College of Computing \\
  Sungkyunkwan University\\%
  Suwon, South Korea \\%
  \texttt{kim.kwangsu@skku.edu}}

\begin{document}

\maketitle
\begin{abstract}
Client contribution evaluation, also known as data valuation, is a crucial approach in federated learning(FL) for client selection and incentive allocation. However, due to restrictions of accessibility of raw data, only limited information such as local weights and local data size of each client is open for quantifying the client contribution. Using data size from available information, we introduce an empirical evaluation method called \textbf{Federated Client Contribution Evaluation through Accuracy Approximation}(FedCCEA). This method builds the Accuracy Approximation Model(AAM), which estimates a simulated test accuracy using inputs of sampled data size and extracts the clients’ data quality and data size to measure client contribution. FedCCEA strengthens some advantages: (1) enablement of data size selection to the clients, (2) feasible evaluation time regardless of the number of clients, and (3) precise estimation in non-IID settings. We demonstrate the superiority of FedCCEA compared to previous methods through several experiments: client contribution distribution, client removal, and robustness test to partial participation.
\end{abstract}

\section{Introduction}
Federated Learning(FL)\citep{federatedlearning_1, mcmahan2017a, federatedlearning_2, openproblems} is an emerging area in the current field of deep learning that unites different models of clients in distributed systems without accessing data. Research on federated learning focuses on various approaches that attempt to reach a similar performance to centralized, optimal models.

One of the approaches is measuring the \emph{client contribution}. An empirical study\citep{notalldataareimportant} suggests that not all data are equally valuable for deep learning models. Discovering high-quality data for frequent usage and filtering low-quality data for elimination are other necessary pre-tasks to attain a highly-performing deep learning model\citep{removing}. Likewise, not all clients are equally important in federated settings; for example, clients with super-biased samples\citep{scaffold, noniid_1} or clients with noisy-labeled samples\citep{overcoming_noisy} may disturb optimization of the federated model. 

The evaluation of client contribution is an indicator of performance improvement in various ways. \textbf{Client selection} \citep{clientselection_1, principledapproach, detectmalicious} is one process that actively uses these estimates, selecting low-contributing clients or low-quality images to exclude them deliberately. Accurate client contribution estimation can noticeably filter malicious or unneeded clients even if countless clients are participating FL. In addition, from an economic perspective, client contribution is a suitable tool for \textbf{incentive allocation} with properties of profit-maximization and fairness\citep{fedgame, learningbasedincentives, collaborativefairness, influence_4}. Every client wants to maximize their profit and concurrently receive an appropriate incentive based on their contribution to the FL. Proper incentive allocation may motivate clients to participate in FL, where the size of the total training samples affects the performance of the global model. 

Then the question is, \emph{"how to evaluate client contribution in FL setting?"} The intuitive contribution measurement is jointly using the data quantity and quality of each client. Data quantity is one of the direct measures non-linearly proportional to the performance\citep{datasize_1, dataquality_1, dataquality_4, learningbasedincentives}. Data quality, which can be evaluated by investigating the overall distribution and performance-affected characteristics of the features and labels\citep{dataquality_2, dataquality_3, pricingprivacy}, complements the nonlinearity of data quantity. Combining data quantity and data quality into one form can act as a proper measure of client contribution. Unfortunately, in a federated learning ecosystem, quantifying the quality of each dataset is an impossible task for the central server because \emph{the blockage of accessing data}\citep{thingstosecure} hinders the analysis of each local dataset. The available local information, which the central server can observe from clients, are local weights(or local gradients) and data quantity used to implement the \emph{FedAvg} algorithm\citep{mcmahan2017a}. This limited local information set is the only ingredient to possibly approximate client contributions. 

Early explorations have suggested a game-theoretic evaluation method, Shapley Value\citep{principledapproach, datashapley}, calculating the marginal test accuracy of the federated model with all possible client subsets including and excluding the specific client. Despite being a theoretically well-structured evaluation method using local model weights, Shapley Value should tackle the exponentially computational burden and imprecise estimation in data heterogeneity environments. More importantly, a business-client FL model requires the clients' choices of data quantity but evaluating every single case of data size selection by Shapley Value is challenging. These weaknesses make the method critically impracticable.




In this work, we introduce a novel approach to client contribution evaluation using data size. Federated Client Contribution Evaluation through Accuracy approximation, also named FedCCEA, learns the data quality of each client by building Accuracy Approximation Model(AAM) with sampled data size. Unlike previous works, FedCCEA provides a more robust and efficient way to approximate the client contribution in practice and allows the clients to participate with selected data size. We demonstrate our strengths by experiments with two public image sets and different data distribution settings. Here are our three main contributions:
\begin{itemize}[leftmargin=0.2in]
    \item We propose a practical client contribution evaluation method that allows partial participation of clients by data size selection.
    \item We evaluates client contribution with a feasible calculation time regardless of the number of clients participating in FL.
    \item We propose a simple accuracy approximation framework, which precisely evaluates client contribution in different types of real-world-like environments such as data heterogeneity and data corruptions.
\end{itemize}

\section{Related Works}
\textbf{Data Valuation} Data Valuation, a similar phrase to Client Contribution Evaluation, is widely studied in recent years to improve centralized machine-learning models and to explain the black-box predictions. Leave-one-out method\citep{influence_2, influence_3} and Influence Function\citep{influence_1, influence_2, influence_3} measure a counterfactual of a single datum and check whether the performance has changed due to the datum. However, these data-counterfactual methods perform poorly; for example, two exactly same but influential points do not value high as they exist together.

Shapley Value\citep{shapley}, a classical concept of cooperative game theory, is on the rise in machine learning to tackle the poor performance of LOO, named Data Shapley\citep{datashapley}. Unlike LOO, Data Shapley compares all possible training data combinations that include a single datum versus combinations that exclude the datum. Moreover, several efficient methods to approximate the actual Shapley value, such as Monte-Carlo SV and gradient-based SV\citep{datashapley} show efforts reducing computational inefficiency that actual Data Shapley suffers. Still, the issue of high computational complexity to data samples remains a weak point for the theory-based valuation methods, as Data Shapley costs $O(2^N)$ of computation complexity for data valuation and Monte-Carlo SV costs $O(N^2\log{N})$.

Empirical methods of data valuation are studied as alternatives to theory-based data valuation. Data Valuation using Reinforcement Learning(DVRL)\citep{dvrl} is the meta-learning framework that jointly learns the data value and trains the primary model by reinforcement learning. This method robustly approximates data values skillfully even in a low-quality dataset or other-domain samples and achieves a high performance of machine-learning tasks by removing parts of low-valued ones.

\textbf{Client Contribution Evaluation for Federated Learning} \emph{FedAvg}\citep{mcmahan2017a}, averaging the updated model parameters, is the most common FL optimization algorithm. Recently, federated learning has altered its focus of analysis on optimization algorithms to data heterogeneity, as the client-drift problem significantly interrupts the optimization of \emph{FedAvg} in non-IID data. Variant algorithms such as FedMA\citep{fedma}, FedProx\citep{fedprox}, and SCAFFOLD\citep{scaffold} are introduced to correct the client-drift problem directly. These optimization methods apply heuristic terms in the aggregation algorithms or average in different ways to stabilize the highly-drifted local gradients far from the optimal point.

Client Contribution Evaluation is another solution of client-drift problem, allowing the measurement of each client’s importance and giving more credits to the most vital clients and fewer credits to the minor clients. Despite the evolution of data valuation methods in centralized machine learning, only a few techniques can be applied in the distributed matters due to the data blockage\citep{thingstosecure}. The DVRL method\citep{dvrl}, the state-of-the-art valuation method, is impossible in FL as it accesses local training data.

The Leave-one-out\citep{influence_4} and the Shapley Value\citep{principledapproach} are applicable valuation methods in distributed systems, which use local parameters or local gradients as a clue of client contribution evaluation. However, even with the introduction of approximated SV methods such as Truncated Monte Carlo SV\citep{TMCshapley}, Group-Testing SV\citep{efficientshapley}, and n-rounds SV\citep{shapley_1}, the game-theoretic methods still suffer from the time-consuming problem. In addition, these two methods do not fully address the imprecision in the data heterogeneity settings and noisy situations.

\begin{figure}[t]
    \centering
    \includegraphics[width = 0.80\textwidth, height = 0.22\textheight]{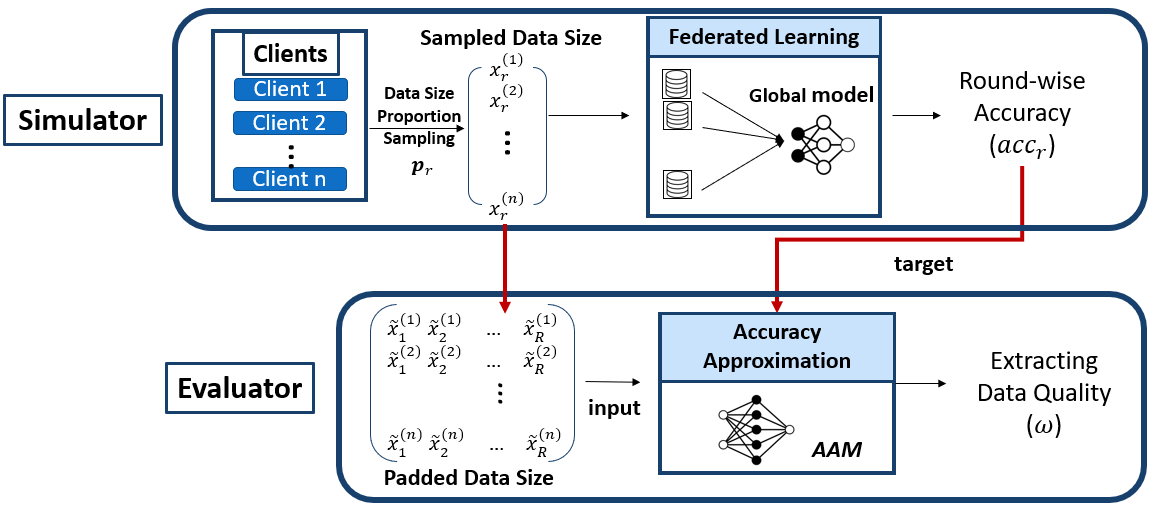}
    \setlength{\belowcaptionskip}{-10pt}
    \caption{Framework of FedCCEA}
    \label{framework}
\end{figure}

\section{Proposed Method}
FedCCEA consists of two phases, the Simulator and the Evaluator. The Simulator is the preparation step of evaluation by simulating FL procedures to obtain the inputs(\emph{sampled data size}) and targets(\emph{round-wise accuracy}) of the Accuracy Approximation Model(AAM) in the Evaluator. After all FL simulations are completed, the Evaluator learns the data quality of each client through AAM.

\subsection{The Simulator}
We denote $n \in \mathbb{N}$, $R \in \mathbb{N}$, and $S \in \mathbb{N}$ respectively as the number of participating clients, the number of rounds per FL simulation, and the number of FL simulations. Moreover, we construct $\mathcal{D}^{(i)}$ of the training data set for each client $i$ and denote $\mid\mathcal{D}^{(i)}\mid$ as the total data size for each client. $\mathcal{D}^{(i)}$ is not used during the testing step; instead, the Simulator uses a separate test set $\mathcal{D}^{t}$ owned by the central server. The main task of the Simulator is implementing FL simulations to obtain the set of \emph{sampled data size}($\mathbf{x}_{r, s}$) and \emph{round-wise accuracy}($acc_{r, s}$). To achieve this goal, we execute the procedure of the Simulator in three steps: Data size Sampling, a single FL iteration, and Testing. The whole routine of these three steps is a single round of a FL simulation, and we repeat $R$-rounds FL simulations $S$ times as initially decided. Algorithm \ref{Simulator} shows the overall procedure of the Simulator phase.

\textbf{Data size Sampling} In the federated learning ecosystem, each client freely selects the size of its local training data for FL in every round. To observe all possible actions of clients, we expand the cases of data size selection by randomly selecting the proportion in the uniform distribution between 0 and 1: $\mathbf{p}_{r,s}$ $= (p^{(1)}_{r,s}, p^{(2)}_{r,s}, ..., p^{(n)}_{r,s}) \sim \mathcal{U}(0, 1)$. We turn  the proportion vector to a real data size vector $\mathbf{d}_{r,s}$ to use in a single FL iteration step. Then, to store data size vector in a normalized term for evaluation, we calculate the standard data size $\mid\mathcal{D}\mid = \frac{\sum{\mid\mathcal{D}^{(i)}\mid}}{n}$ and determine a scaled data size vector $\mathbf{x}_{r,s}$:

\begin{gather}
    \mathbf{d}_{r,s} = (d^{(1)}_{r,s}, d^{(2)}_{r,s}, ..., d^{(n)}_{r,s}) \quad \text{where } d^{(i)}_{r,s} = \mid\mathcal{D}^{(i)}\mid \times p^{(i)}_{r,s}, \quad\quad \mathbf{x}_{r,s} = \frac{\mathbf{d}_{r,s}}{\mid\mathcal{D}\mid}
\end{gather}

Note that this scaled data size vector $\mathbf{x}_{r,s}$, also defined as \emph{sampled data size}, accelerates the convergence of AAM and allows to compare the data size between clients while learning the client contribution in the Evaluator phase.

\textbf{A single FL iteration} The next step is a one-epoch FL classification task. The global model is a DNN model such as MLP or CNN, as assumed here. During this step, the central server renews the global model parameter $\theta^{G}_{r}$ based on the \emph{sampled data size} of clients. Each client updates their local model weights $\theta^{(i)}_{r}$ using only $d^{(i)}_{r,s}$ of their dataset, which is their actual size to train in this round. Then, the central server aggregates the local model weights by \emph{FedAvg} algorithm as usual. Based on the obtained information from the clients, \emph{FedAvg} is reformulated as $\theta^{G}_{r} = \sum_{i=1}^{n}{\frac{d^{(i)}_{r,s}}{\sum_j{d^{(j)}_{r,s}}}}\times\theta^{(i)}_{r}$.

\begin{algorithm}[t]
\SetAlgoLined
\KwIn{Number of clients $n$, number of rounds per FL simulation $R$, number of simulations $S$, training set $\mathcal{D}^{(i)}$ for each client $i=1, 2, ..., n$, shared test set $\mathcal{D}^t$, and standard data size $\mid\mathcal{D}\mid = \frac{\sum{\mid\mathcal{D}^{(i)}\mid}}{n}$}
\SetKwProg{Clients}{Clients Execute}{:}{end}
\BlankLine
Initialize empty list $E$\;
\For {$s=1,2,...,S$}{
 Initialize parameters of the global model $\theta^{G}_{0}$ \;
 \For {$r=1,2,...,R$}{
 Sample a data size proportion vector $\mathbf{p}_{r,s}$ $= (p^{(1)}_{r,s}, p^{(2)}_{r,s}, ..., p^{(n)}_{r,s}) \sim \mathcal{U}(0, 1)$\;
 Derive a real data size vector $\mathbf{d}_{r,s}$ = $(d^{(1)}_{r,s}, d^{(2)}_{r,s}, ..., d^{(n)}_{r,s})$ where $d^{(i)}_{r,s} = \mid\mathcal{D}^{(i)}\mid \times p^{(i)}_{r,s}$\;
 Derive a scaled data size vector $\mathbf{x}_{r,s}$ = $\frac{\mathbf{d}_{r,s}}{\mid\mathcal{D}\mid}$\;
 \Clients{}{
 Collect global parameters $\theta^{G}_{r-1}$\;
 Update local models using $\mathbf{d}_{r,s}$ and obtain $\theta^{(i)}_r$ for each client $i$\;}
 Collect $\theta_{r}$ = [$\theta^{(1)}_r, ..., \theta^{(n)}_r$] from all clients\;
 Implement \emph{FedAvg} Algorithm and update $\theta^{G}_{r}$\;
 Test the updated global model using $\mathcal{D}^t$ and obtain a round-wise test accuracy $acc_{r,s}$\;
 Store ($\mathbf{x}_{r,s}$, $acc_{r,s}$) into $E$\;}}
\BlankLine
\KwRet{list $E$}
\caption{FedCCEA - Procedure of the Simulator}\label{Simulator}
\setlength{\belowcaptionskip}{-10pt}
\end{algorithm}


\textbf{Testing} After a single FL iteration, the central server tests the advanced global model with a separate test set $\mathcal{D}^{t}$ to gain a scalar value of test accuracy in this round($acc_{r,s}$), defined as \emph{round-wise accuracy}. Then, the \emph{sampled data size}($\mathbf{x}_{r,s}$) and \emph{round-wise accuracy}($acc_{r,s}$) are stored for the usage in the Evaluator phase.

\subsection{The Evaluator} \label{evaluator}

The Evaluator phase is a separate process from the Simulator phase but takes a substantial role in extracting the data quality from a learned Accuracy Approximation Model. This phase begins once the Simulator phase is completed so that the Evaluator can obtain whole stored results from the Simulator. The overall procedure of the Evaluator phase is enumerated in Algorithm \ref{Evaluator}.

\subsubsection{Model Structure of AAM}

\begin{figure}[ht]
    \centering
    \subfigure[The First Layer of the AAM]
    {\centering
    \includegraphics[width= 0.33\textwidth, height = 0.13\textheight]{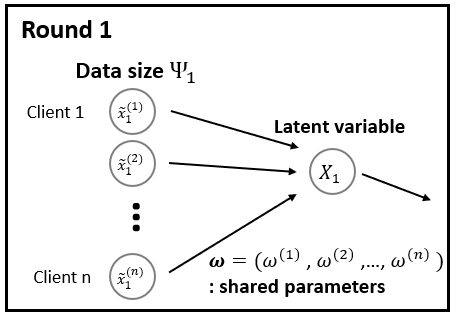}
    \label{first_layer}}
    \qquad
    \subfigure[Overall Architecture of the AAM]
    {\centering
    \includegraphics[width= 0.5\textwidth, height = 0.13\textheight]{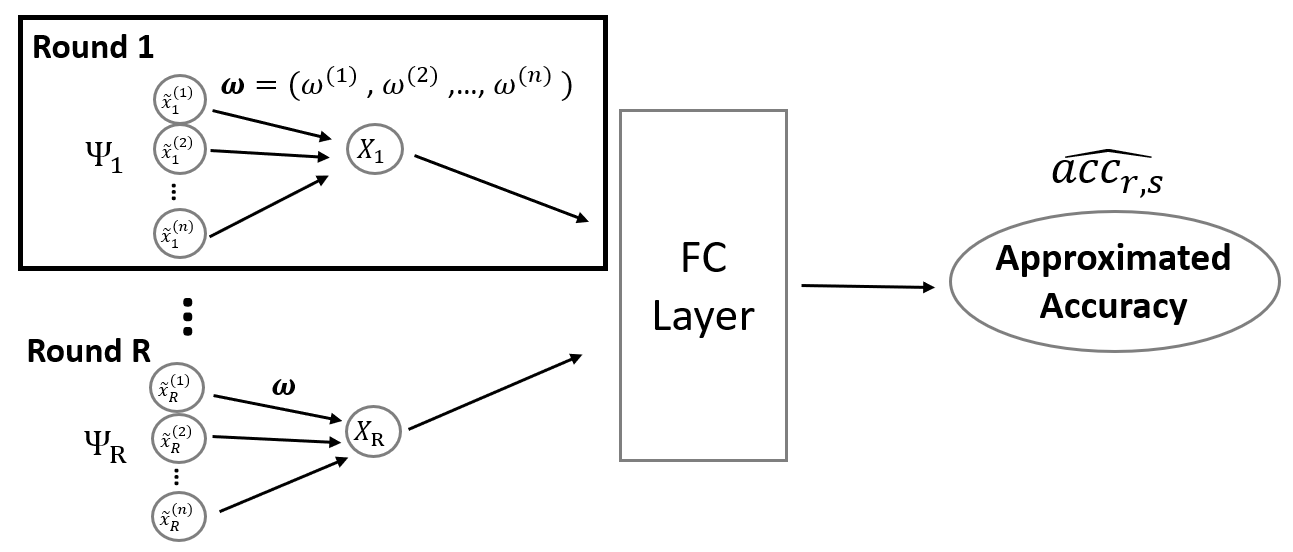}
    \label{aam_structure}}
    \caption{Model Structure of the AAM}
    \label{aam}
\end{figure}
The Accuracy Approximation Model(AAM) is a regression model, approximating the \emph{round-wise accuracy} by the \emph{sampled data size}. This model demonstrates the impact of the \emph{sampled data size} utilized in each round on the \emph{round-wise accuracy}. The uniqueness of the model is that \emph{sampled data size} in previous rounds also affects \emph{round-wise accuracy}; for example, accuracy in round $r$ is also affected by data size set in round 1 to round $r-1$.

\begin{wrapfigure}{h}{0.33\textwidth}
\centering
\includegraphics[width=0.33\textwidth]{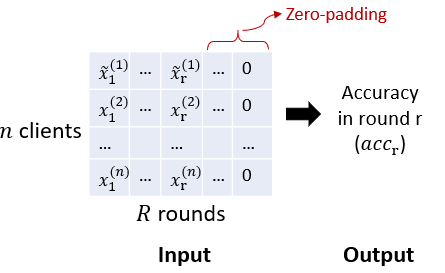}
\caption{\textbf{Construction of input vector set($\Psi_r$) for AAM}. The elements of $\Psi_{r} \in \mathbb{R}^{n \times R}$ that are over round $r$ are zero-padded.}
\label{input}
\end{wrapfigure}

In this regard\footnote{RNN-based and Reinforcement Learning-based models are considerable, but they require excessively complex frameworks and need dynamic window sizes, which are not appropriate for this situation.}, we design the AAM in a simple Multi-Layer Perceptrons(MLP) framework with several distinctive tools, forming $g: \mathbb{R}^{n \times R} \longrightarrow [0,1]$ shown in Figure \ref{aam_structure}. This model allows the stored data to be considered in sequential matters by organizing \textbf{the static-sized inputs with zero-padding} of the future actions. Also, it enables the measurement of client contribution by \textbf{shared weights}.

An input vector of the AAM, $\Psi_{r} \in \mathbb{R}^{n \times R}$, is constructed with the experienced \emph{sampled data size} sets. From static $n \times R$ shape, we list the \emph{sampled data size} sets before round $r$. Then, we \textbf{zero-pad} the rest of the space, which are the upcoming actions after round $r$. For each sample, we can assume as proceeding all R rounds of a single FL simulation, including the actions of all clients not participating FL after round $r$. Figure \ref{input} demonstrates the construction of zero-padded input vectors.

The focal point of the architecture is on the first layer of AAM, where it contains \textbf{shared weights} $\omega \in \mathbb{R}^{n}$. Shared weights are widely used in CNN architecture as a shape of convolution filters that extract the local features of the image data. Similarly, the first layer of the AMM is structured with round-wise shared weights to extract the importance of data size for each client.

The inputs are split into $R$ round-wise sets, expressed as $\Psi_r = (\tilde{x}_r^{(1)}, \tilde{x}_r^{(2)}, ... , \tilde{x}_r^{(n)})$ for each set. The model initially designs a linear regression for each round-wise set as $X_r(\Psi_r; \omega) = \tilde{x}_r^{(1)}\omega^{(1)}+\tilde{x}_r^{(2)}\omega^{(2)}+...+\tilde{x}_r^{(n)}\omega^{(n)}$ using the shared parameter vector $\omega = (\omega^{(1)}, \omega^{(2)}, ..., \omega^{(n)})$. $X_r$ may represent the expected total value of the given data size set in round $r$.

The remaining layers of the AAM are constituted as fully connected layers(FC Layers), $f: X \longrightarrow [0,1]$, that return a single approximated accuracy. Concatenated vector $X = (X_1, X_2, ..., X_R)$, originated from the linear regressions $X_r(\Psi_r; \omega)$, is the input of these layers. These layers may closely be related to the impact of each round; in common sense, earlier rounds are more important than later rounds. Thus, we formulate the corresponding optimization problem of the AAM as:

\begin{equation}
\label{optimization_aam}
\begin{split}
    \min_{\mathbf{\Omega}}{\mathcal{L}(g(\mathbf{\Psi}; \mathbf{\Omega}), acc)} &= \min_{\mathbf{\omega}, \mathbf{\Omega}_{-\omega}}{\mathcal{L}(f((X_1(\Psi_1; \mathbf{\omega}),\dots,X_R(\Psi_R;\mathbf{\omega})); \mathbf{\Omega}_{-\omega}), acc)}
\end{split}
\end{equation}
Note that $\omega \in \mathbf{\Omega}$ refers to the shared weight vector in the first layer, and $\mathbf{\Omega}_{-\omega} \in \mathbf{\Omega}$ refers to the remaining weight vectors in the FC layers.

\begin{algorithm}[t]
\SetAlgoLined
\KwIn{Number of global rounds $R$, number of simulations $S$, and results of the Simulator $E = \{(\mathbf{x}_{r,s}, acc_{r,s})\}_{r=1,...,R, s=1,...,S}$}
\BlankLine
Initialize parameters $\mathbf{\Omega}$ of AMM\;
\For {$s=1,2,...,S$}{
    \For {$r = 1,2,...,R$}{
    Construct an input vector $\Psi_{r,s} \in \mathbb{R}^{n\times R}$\;
    List all data size set $\mathbf{x}_{r,s}$ under round $r$ in $\Psi_{r,s}$ and zero-pad for the rests\;}}

Collect $\Psi = \{\Psi_{r,s}\}_{r=1,...,R, s=1,...,S}$ and $acc = \{acc_{r,s}\}_{r=1,...,R, s=1,...,S}$\;
\While {until convergence}
    {Using $\Psi$ inputs and $acc$ targets, optimize $\mathbf{\Omega}$ from AMM : $g(\Psi;\mathbf{\Omega})$\;}

Extract the shared weights($\mathbf{\omega}$) of the first layer\;
\BlankLine
\KwRet{weight vector $\mathbf{\omega}$}
\caption{FedCCEA - Procedure of the Evaluator}\label{Evaluator}
\end{algorithm}

\subsubsection{Client Contribution Estimation}
Reminding that data quality is the main index to seek the estimation of client contribution, we extract the data quality of each client through the AAM. In AAM, the shared weight vector($\omega$) represents the importance of the data size set to the latent variables $Xs$, representing the round-wise impact. In contrast, the weights of the FC layers($\mathbf{\Omega}_{-\omega}$) focus on the impact of rounds. Regardless of the current round state, we can interpret the $\omega$ as averaged data quality of each client. Given data size set $x = (x^{(1)}, x^{(2)}, ... , x^{(n)})$, we can quantify the client contribution by the multiplication of data quality and quantity: $Client\; Contribution\; Value(v_i) = x^{(i)} \times \omega^{(i)}$.

\section{Experiments}
\subsection{Experiment Setups}\label{experiment_setups}

We verify the suitability of the measured client contributions of FedCCEA on two public image sets\footnote{MNIST consists of 60,000 training images and 10,000 test images with 10 nearly-balanced labels, and EMNIST \emph{by class} consists of 731,668 training images and 82,587 test images with 62 unbalanced labels.}(MNIST\citep{mnist}  \& EMNIST \emph{by class}\citep{emnist}) and different data distribution settings. We compose three different types of tests to show the strengths of our method from different angles.

\textbf{Baseline methods} We compare the results with two baseline methods that are applicable in the distributed settings: \textbf{Leave-one-out method}(LOO)\citep{influence_4} and \textbf{Truncated Monte-Carlo Shapley Value method}(TMC-SV)\citep{principledapproach}, the approximated estimation of Shapley Value.

\textbf{Federated Learning settings} We use the three-layers MLP for MNIST and two-layers MLP for EMNIST as the federated model. The number of clients($n$) and the number of global rounds($R$) are fixed to $(20, 50)$ for all three methods. For FedCCEA, we additionally set the number of FL simulations($S$) to 100.

Also, the epochs of local training($L$), batch size($B$), and learning rate($lr$) of updating local models are essential hyper-parameters that influence test accuracy and training time in FL. We assume ($L, B, lr$) to be set as $(3, 32, 0.001)$ for MNIST and $(1, 256, 0.05)$ for EMNIST.

\textbf{Data distribution settings}\footnote{The illustrations and the results of different types of data distribution settings that are not mentioned in this section are left in Appendix.} The degree of difficulty for client contribution evaluation surges as the degree of data heterogeneity rises. We show the stability of our method by pre-defined three different types of data distribution settings:
\begin{itemize}[leftmargin=0.2in]
    \item IID: All 20 clients equally contain 10 classes for MNIST and 62 classes for EMNIST.
    \item Weak non-IID: Each client contains 5 classes for MNIST and 30 classes for EMNIST.
    \item Strong non-IID: Each client contains 2 classes for MNIST and 5 classes for EMNIST.
\end{itemize}

In addition, noise is another exogenous factor that interferes with the accurate evaluation of client contribution. Noise-injected clients confuse the contribution of clean clients, as the importance of clean clients may change due to the significant parameter variation for noise-injected clients. Thus, we provide different noise settings to show our superiority even in the real-world-like environments:
\begin{itemize}[leftmargin=0.2in]
    \item No noise: Usual data distribution setting with no noise
    \item Label noise: Following \citep{labelnoise1}, select 20\% of clients and give a label noise to 40\% proportion of the samples; for example, change label `$7$' to `$5$'.\footnote{We also diversify noise settings: injecting label noise to 10\% and 20\% of the samples and giving pattern-backdoor attacks\citep{samplenoise1} to the 20\% of the samples. The experiments of these settings are recorded in Appendix.}
\end{itemize}


\textbf{Client Contribution Index} Since three evaluation methods extract client contribution values in different ranges, we standardize these values into one unified index called \emph{Client Contribution Index}(CCI). CCI is newly measured by calculating the relative importance between the clients, ranging $[0,1]$. The negative values, which are out of the range for measuring CCI, are initially set to 0, meaning zero contribution to the federated model.\footnote{Despite valuing 0 to negative contributors, we leave the rank of contribution between clients for the client removal experiment in Section \ref{Client Removal}.} By denoting $v_i$ as the value of client contribution measured in a given evaluation method, we calculate CCIs as:

\begin{equation}
\begin{split}
\label{CCI}
    Client\;Contribution\;Index(CCI) = \frac{\tilde{v}_i}{\sum_j{\tilde{v}_j}}
    \quad\quad
    \text{where }
    \tilde{v}_i  = 
    \begin{cases}
    0 & \text{if $v_i\leq 0$,}\\
    v_i & \text{otherwise.}
    \end{cases}
\end{split}
\end{equation}

\subsection{Client Contribution Skewness}
\label{Client Contribution Skewness}

This experiment aims to measure CCI for each client and checks the CCI distribution whether it is nearly uniformed even in the strong non-IID setting. Ideally, different data distribution may generate uneven contributions between participants. However, the least contributed client still takes a minor part in the federated model unless it is a malicious client containing corrupted samples.

We use three evaluation methods with no noise. Since negative contributions exist when noise is injected to several clients, and judging the suitability of negative contributions is challenging, we exclude noise injection setting in this experiment.

\begin{figure}[ht]
    \centering
    \subfigure[MNIST]
    {\centering
    \includegraphics[width= 0.9\textwidth, height = 0.09\textheight]{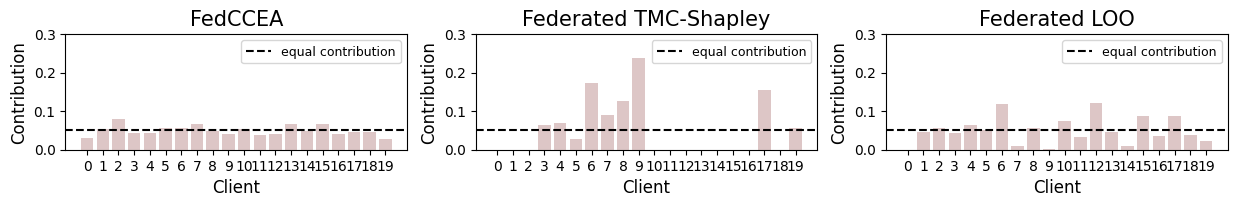}
    \label{MNIST_contribution}}
    
    \subfigure[EMNIST]
    {\centering
    \includegraphics[width= 0.9\textwidth, height = 0.09\textheight]{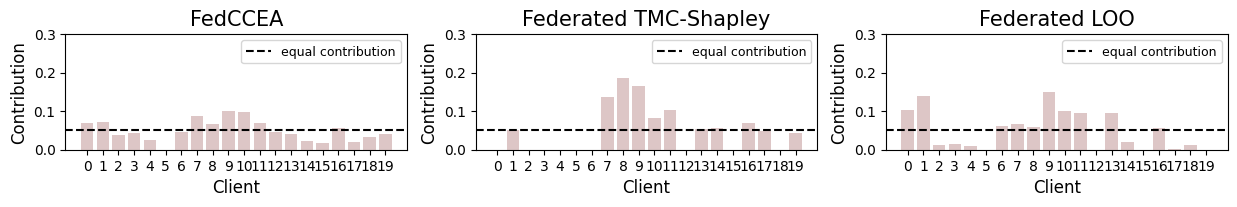}
    \label{EMNIST_contribution}}
    \caption{Client Contribution Index(CCI) in strong non-IID setting of three evaluation methods, FedCCEA(left), TMC-Shapley(center), and LOO(right). (a) uses MNIST data while (b) uses EMNIST data. The horizontal dashed line represents the equal contributions valuing 0.05 for all clients.}
    \label{noiseinjection}
\end{figure}

As the results shown in Figure \ref{noiseinjection}, two theoretic-based methods appear to have high skewness of contribution, concentrating on few high contributors and severely giving values to others. TMC-SV evaluates zero CCIs to 11 clients for MNIST and nine clients for EMNIST, while LOO evaluates zero CCIs to one client for MNIST and four clients for EMNIST. The client-counterfactual methods consider specific subsets or specific rounds that record highly negative values of marginal accuracy. These negative values offset the positive contribution cases and remain negative on average; in other words, we say strong losses overwhelm weak gains. On the contrary, FedCCEA shows the most equally distributed contribution between clients. The minor contributors for MNIST and EMNIST value 0.0285 and 0.0001, respectively.

\begin{table}[t]
  \centering
  \begin{tabular}{p{0.25\textwidth}l >{\centering}p{0.15\textwidth}l >{\centering}p{0.15\textwidth}l >{\centering}p{0.15\textwidth}l >{\centering}p{0.15\textwidth}l}
    \toprule
    \multicolumn{2}{c}{} &
    \multicolumn{3}{c}{Client Contribution Evaluation Method} \tabularnewline
    \cmidrule(r){3-5}
    Dataset & Base & FedCCEA & TMC-SV & LOO \tabularnewline
    \midrule
     MNIST & 85.60\%  & \textbf{85.60\%} & 74.29\% & 85.29\% \tabularnewline
     EMNIST & 59.64\%  & \textbf{59.64\%} & 50.52\% & 54.65\% \tabularnewline
    \bottomrule
  \end{tabular}
  \caption{\textbf{Exclusion of zero contributors}: Test accuracy of federated learning after excluding clients with zero CCIs. Base is the test accuracy of federated learning with all 20 clients participating. The best results among three client contribution evaluation methods are shown in \textbf{bold}.}
  \label{retrain_contribution}
\end{table}

Though two baseline methods define several clients as zero contributors, these clients still contribute to the federated model's performance improvement. Table \ref{retrain_contribution} shows the retrained model's test accuracy, excluding clients with zero CCIs evaluated by each method. While FedCCEA results in the same performance as Base since no clients are measured zero CCIs, the other two ways have to experience a significant decline of performance due to the exclusion of incorrectly-measured zero contributors.

\subsection{Client Removal}\label{Client Removal}
In this experiment, we remove clients in both descending and ascending order of CCIs to investigate the correctness of the contribution tier that three evaluation methods measured. If the client contribution is precisely estimated, removing clients with low CCIs in ascending order will consistently retain high test accuracy. In contrast, the performance of eliminating the highest contributors will drop substantially as incrementally removing clients.

\begin{figure}[ht]
    \centering
    \subfigure[MNIST]
    {\centering
    \includegraphics[width= 0.46\textwidth, height =0.2\textheight]{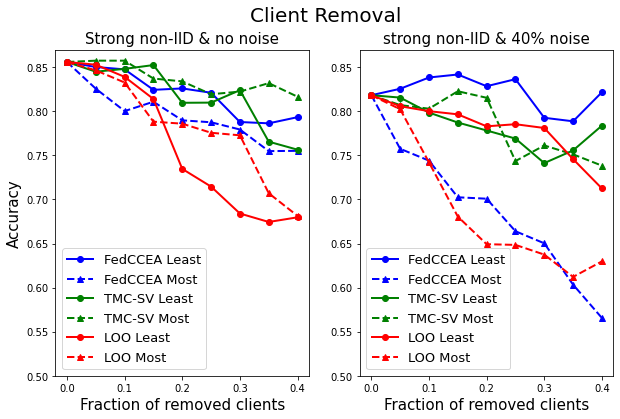}
    \label{MNIST_removal}}
    \qquad
    \subfigure[EMNIST]
    {\centering
    \includegraphics[width= 0.46\textwidth, height =0.2\textheight]{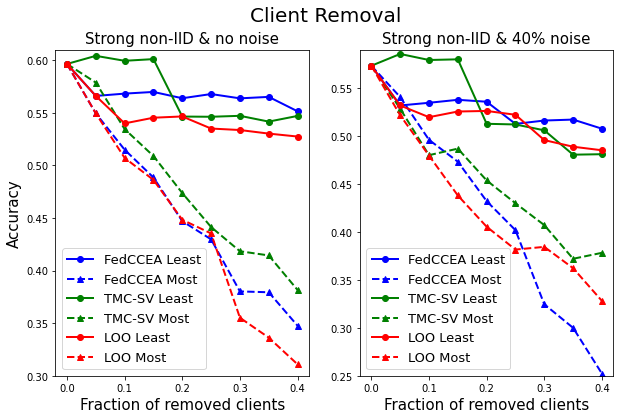}
    \label{EMNIST_removal}}
    \caption{Test accuracy after removing the lowest(straight line) and highest(dashed line) contributors in order using (a)MNIST and (b)EMNIST in the strong non-IID settings. The left graph shows the environment without noise injection, and the right graph shows the environment with 40\% label noise. Ideally, the straight line should maintain high performance while the dashed line should drop dramatically after removing clients.}
    \label{removal}
    
\end{figure}
As shown in Figure \ref{removal}, FedCCEA(blue) reveals consistencies of the performance until the 40\% proportion of lowest contributors are excluded. Moreover, FedCCEA experiences a significant decline in performance as the highest contributors are removed sequentially, especially with 40\% label noise settings for both MNIST and EMNIST. On the other hand, TMC-SV and LOO for MNIST experience a reversal between removing the highest and lowest contributors. This performance reversal of high contributors and low contributors can indicate a flawed evaluation method. Thus, we doubt that these baseline methods precisely evaluate client contribution.\footnote{TMC-SV for EMNIST shows the highest performance during $0 \sim 0.15$ proportion of removing the lowest contributors. Nevertheless, the `TMC-SV Least' does not result consistent performance after extra removal, and more importantly, the performance of `TMC-SV Most' does not critically decline as much as FedCCEA.}

\subsection{Robustness of Client Removal to Partial Participation}\label{robustness}
The distinctiveness of FedCCEA compared to previous works is that clients are allowed to choose how much data they use for federated learning in every round. The amount of data they choose substantially varies the results of \emph{FedAvg}, which eventually affects the test accuracy. Thus, an accurate client contribution approximation for partial participation is needed as well.

This experiment aims to prove the robustness to partial participation by testing the same experiment in section \ref{Client Removal}. We (1) randomly assign the data size of each client in every round, (2) rank the CCIs in both descending and ascending orders, and (3) retrain the federated model by removing a given proportion of highest and lowest contributors in every round. We compare the results with the case of FedCCEA using full data size.\footnote{Two baseline methods are impossible to apply the partial participation of clients since extra evaluations are needed for every action, so we only consider our method's results.}

\begin{figure}[ht]
    \centering
    \subfigure[MNIST]
    {\centering
    \includegraphics[width= 0.46\textwidth, height =0.2\textheight]{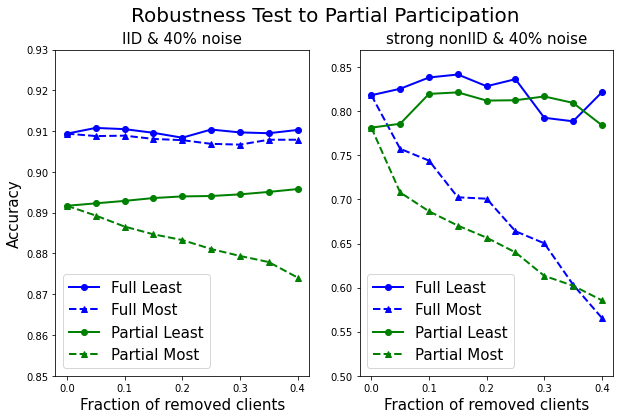}
    \label{MNIST_partial_participation}}
    \qquad
    \subfigure[EMNIST]
    {\centering
    \includegraphics[width= 0.46\textwidth, height =0.2\textheight]{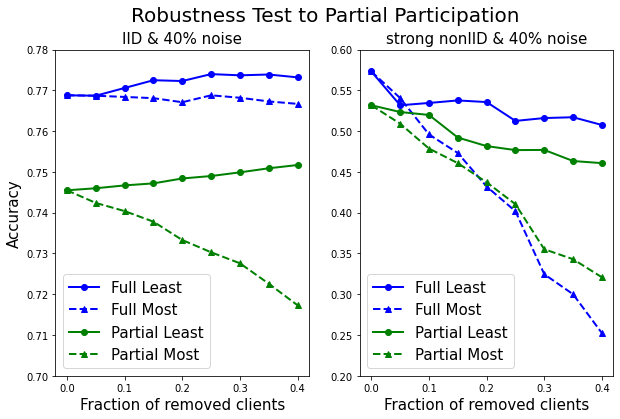}
    \label{EMNIST_partial_participation}}
    \caption{FedCCEA Robustness Test of Client Removal to Partial Participation with 40\% noise injection using (a) MNIST and (b) EMNIST. The left graph shows the results in IID setting while the right graph shows the results in strong non-IID setting.}
    \label{partial_participation}
\end{figure}

As shown in Figure \ref{partial_participation}, both datasets show the outstanding performance of client removal even in the situation of partial participation. The data deficiency from partial participation(green) causes the overall accuracy decline compared to full participation(blue). Still, partial participation has a wider gap between `Least' and `Most' than full participation in IID-setting. We can interpret that FedCCEA explicitly considers the impact of low data size by filtering the useless clients who use low data quantity and accurately ranks the contribution of clients regarding the data size. Not only in the IID setting, FedCCEA also correctly evaluates the client contribution in the non-IID environments, resulting in a consistent performance like full participation.

\section{Conclusion}
In this paper, we proposed FedCCEA, an empirical method of client contribution evaluation without accessing raw data. Practicality, feasible evaluation time, and consistency within diverse settings make this method more powerful than previous methods. We prove these contributions with three different experiments. Of course, some datasets with complex FL tasks may not accurately estimate client contributions in specific settings; but the problem mainly comes from the inadequacy of FL optimization algorithms. We should handle this case by applying alternative algorithms to achieve high performance in the first place and then operating superior evaluation methods for correct client selection and incentive allocation.


\bibliographystyle{abbrv}
\setcitestyle{year, open{[}, close={]}}
\bibliography{reference}



\begin{appendices}
\section{Details of the Evaluator}
In Section \ref{evaluator}, we explained the structure of the Accuracy Approximation Model with two specific tools: zero-padded inputs and shared weights of the first layer. In more detail, we arrange ten nodes in the first FC layer for all experimental settings. For the simplicity of CCI calculation, we put the weight constraints for negative values of weights in the first layer to 0 so that the shared weights, known as data quality, are measured as 0 for minus contributors. Moreover, the sigmoid activation function is added in the first layer of the AAM, while the remaining layers are constructed with linear activation functions. During the training stage of the AAM, we set the learning rate differently regarding the difficulty of the data.\footnote{Learning rate : 0.01 to MNIST results and 0.05 to EMNIST results.}

\begin{figure}[ht]
    \centering
    \subfigure[MNIST \& IID with no noise]
    {\centering
    \includegraphics[width= 0.42\textwidth, height = 0.18\textheight]{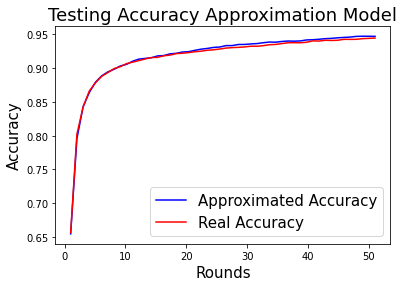}
    \label{mnist_aam}
    }
    \qquad
    \subfigure[EMNIST \& IID with no noise]
    {\centering
    \includegraphics[width= 0.42\textwidth, height = 0.18\textheight]{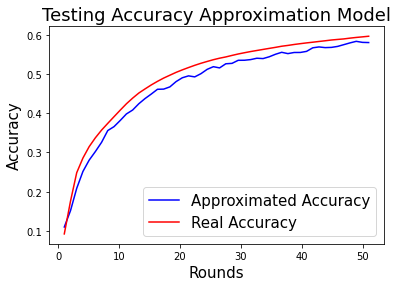}
    \label{emnist_aam}
    }
    \caption{The difference between approximated accuracy from the AAM and the real test accuracy.}  
    \label{aam_result}
\end{figure}

As more results from the Simulator are collected, the AAM can estimate a more precise approximation of test accuracy. Figure \ref{aam_result} reports the difference between approximated accuracy and the actual test accuracy, with the hyper-parameters initially defined S=100 and R=50. The trivial difference for both datasets in Figure \ref{aam_result} justifies the interpretation of the shared weights that affects the test accuracy.

\section{Extra Experiment Setups}
\subsection{Data Distribution Settings}
\begin{landscape}
\begin{figure}[ht]
    \centering
    \subfigure[IID Setting]
    {\centering
    \includegraphics[width= 1.5\textwidth, height = 0.26\textheight]{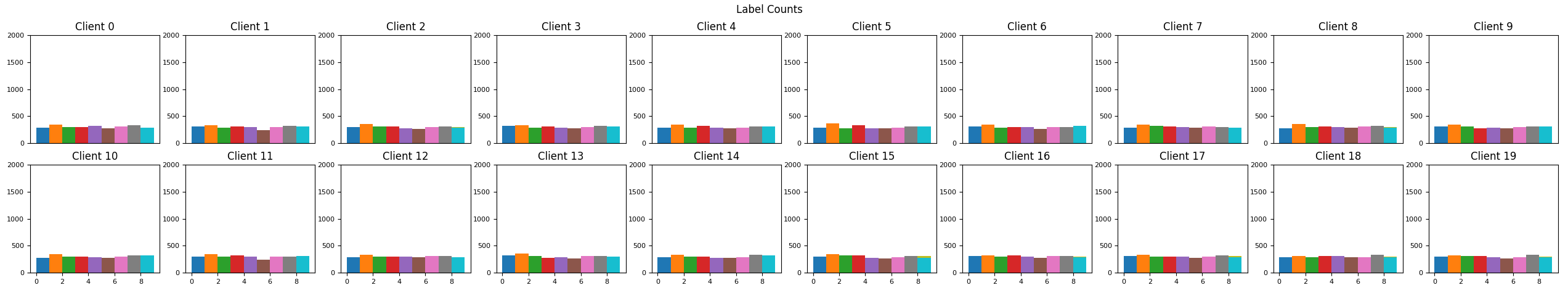}
    \label{mnist_iid}}
    
    \subfigure[Weak non-IID Setting]
    {\centering
    \includegraphics[width= 1.5\textwidth, height = 0.26\textheight]{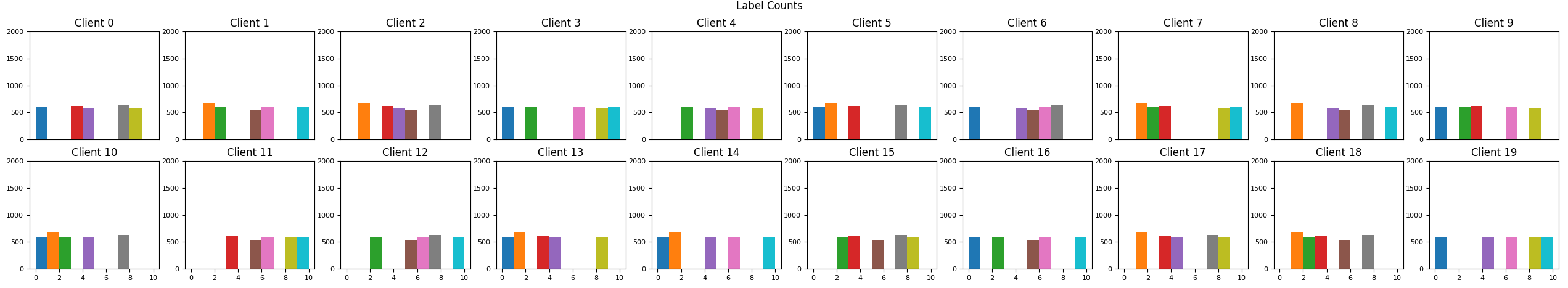}
    \label{mnist_noniid5}}
    
    \subfigure[Strong non-IID Setting]
    {\centering
    \includegraphics[width= 1.5\textwidth, height = 0.26\textheight]{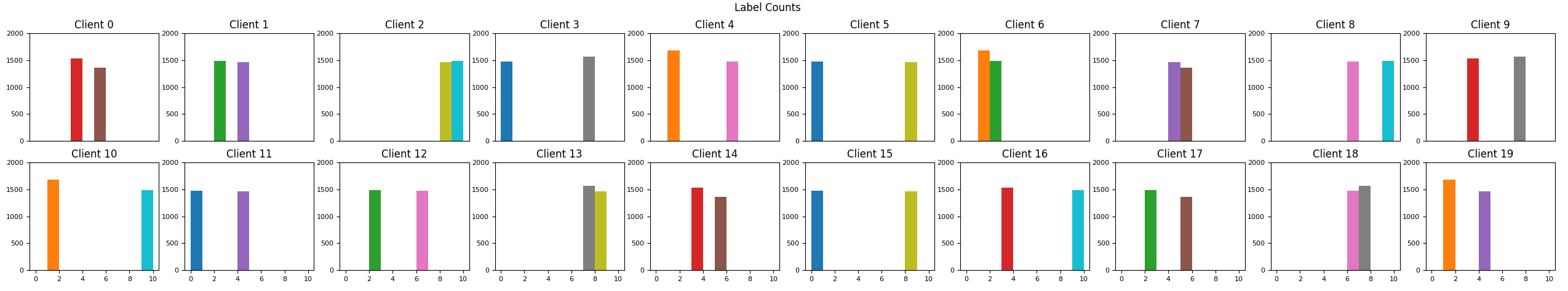}
    \label{mnist_noniid2}}
    \caption{\textbf{Three Data Distribution Settings for MNIST} X-axis:index of labels(10 labels), Y-axis: data counts for each label} 
    \label{data_distribution_settings_MNIST}
\end{figure}
\end{landscape}

\begin{landscape}
\begin{figure}[ht]
    \centering
    \subfigure[IID Setting]
    {\centering
    \includegraphics[width= 1.5\textwidth, height = 0.26\textheight]{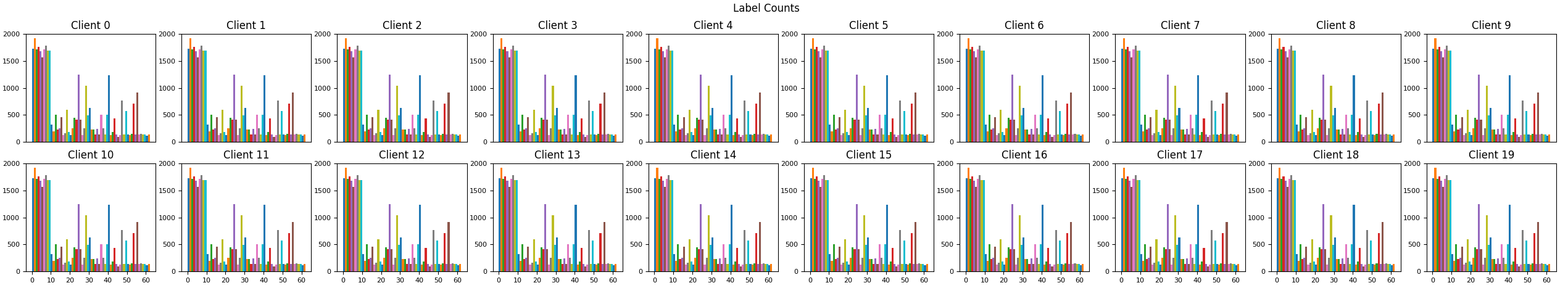}
    \label{emnist_iid}}
    
    \subfigure[Weak non-IID Setting]
    {\centering
    \includegraphics[width= 1.5\textwidth, height = 0.26\textheight]{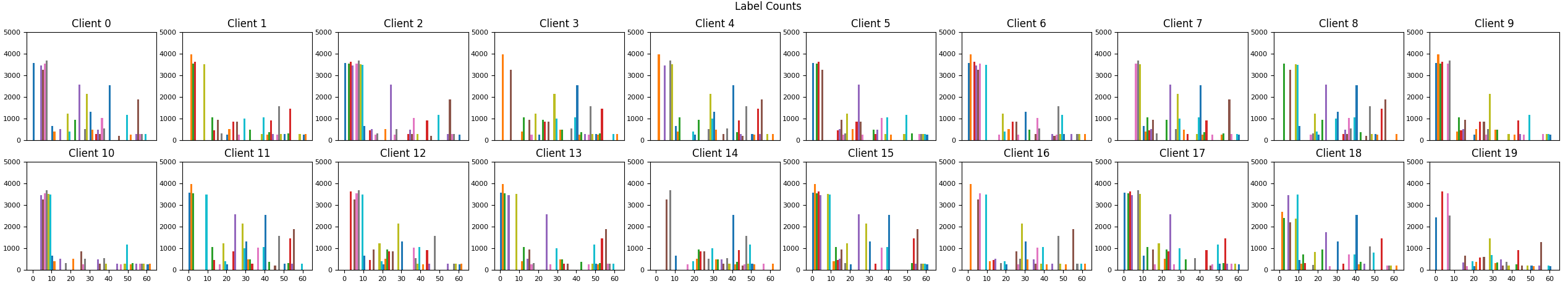}
    \label{emnist_noniid30}}
    
    \subfigure[Strong non-IID Setting]
    {\centering
    \includegraphics[width= 1.5\textwidth, height = 0.26\textheight]{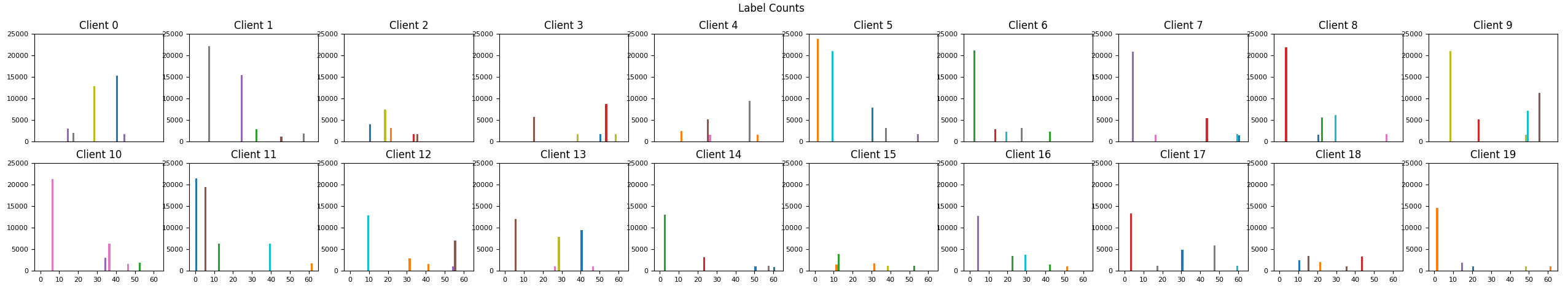}
    \label{emnist_noniid5}}
    \caption{\textbf{Three Data Distribution Settings for EMNIST} X-axis:index of labels(62 labels), Y-axis: data counts for each label}
    \label{data_distribution_settings_EMNIST}
\end{figure}
\end{landscape}

As mentioned in Section \ref{experiment_setups}, we divide the training data distribution settings into three types by the degree of data heterogeneity. The IID setting supports all clients containing all classes with equal proportions of total training samples. The plots in Figure \ref{mnist_iid} and \ref{emnist_iid} demonstrate that all clients have equal distribution of data with other clients. On the other hand, the data of each client concentrate on specific classes in non-IID settings. While clients in weak non-IID settings have half of the classes, only few labels are distributed to each client in strong non-IID settings. The non-IID settings for MNIST and EMNIST are illustrated in Figure \ref{mnist_noniid5}, \ref{mnist_noniid2} and \ref{emnist_noniid30}, \ref{emnist_noniid5}.

To directly compare the results with two baseline methods in the experiments, the data distribution settings and the order of samples should not be altered during the evaluation steps and the experiments. Moreover, the measured client contribution values are only meaningful in the distribution settings used for evaluation. For these reasons, once defining and initializing a distribution setting, we do not renew the environment again; in other words, the settings in Figure \ref{data_distribution_settings_MNIST} and \ref{data_distribution_settings_EMNIST} are fixed during the evaluation and experiments. Also, the samples of each client is not shuffled again once it was initially settled to construct identical batch for every simulation and test.

\subsection{Noise Injection Settings}
\begin{figure}[t]
    \centering
    \subfigure[A Noisy Client with 40\% Label Noise]
    {\centering
    \includegraphics[width= 0.55\textwidth, height = 0.20\textheight]{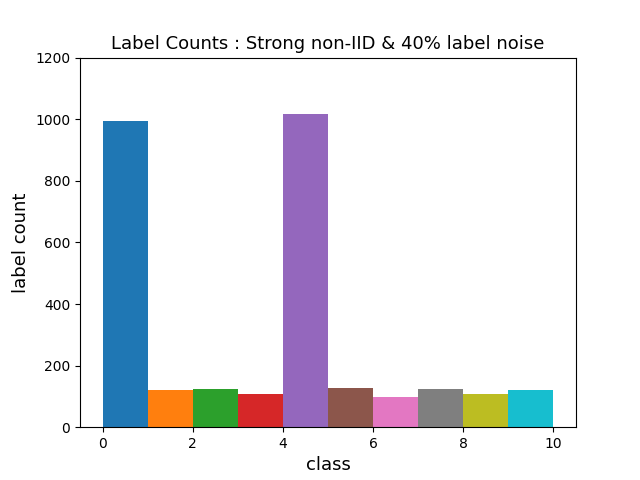}
    \label{Label_Noise}}
    \qquad
    \subfigure[Example of Sample Noise]
    {\centering
    \includegraphics[width= 0.38\textwidth, height = 0.20\textheight]{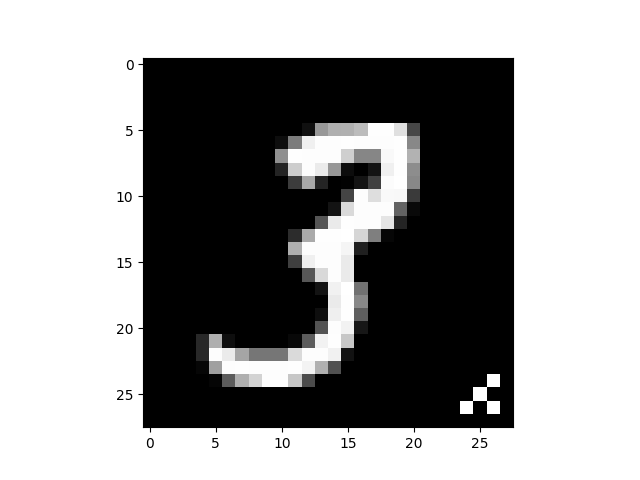}
    \label{Sample Noise}}
    \caption{Illustrations of Noise Injection for MNIST}  
    \label{noiseinjection_example}
\end{figure}

To verify the strengths of FedCCEA, we inject a label noise to 40\% of training samples in 20\% of clients in pre-defined data distribution settings. Figure \ref{Label_Noise} illustrates a noisy client in the strong non-IID, MNIST setting that contains 40\% label noise; for instance, change label `$0$' to `$3$'. The dataset without label noise would only contain class $\{0,4\}$, but the other classes are present as the label noise is injected into the noisy clients. We design the noise settings for EMNIST in the same way.
\footnote{Datasets in federated settings and simulation results are stored in: \\ \url{https://drive.google.com/drive/folders/1CtnhoMUq9PBX7YY97Qgxm6P_4exVc_u7?usp=sharing}}

In addition to 40\% label noise injection, different noise environments are designed to diversify the cases. We vary the extent of label noise injected to noisy clients: 20\% label noise and 10\% label noise to 20\% of participants. Moreover, as shown in Figure \ref{Sample Noise}, we also injected sample noise as pattern-backdoor attacks, giving a pattern of white pixels in the bottom right corner of the selected images. This type of noise may interrupt the optimization of federated learning but not as crucial as label noise. In summary, here are the noise injection settings we initially design: No Noise, 40\% Label Noise, 20\% Label Noise, 10\% Label Noise, and 20\% Sample Noise.

\clearpage

\section{Extra Experiment Results}
\subsection{Evaluation Time}

We check the evaluation time for each method running on Intel Xeon Processor(Skylake, IBRS) with 16 CPU cores. Each client occupies a single CPU core and synchronously train each local dataset with its occupied core. We do not use a single GPU core in our FL framework. Within our resource, we check the evaluation time for different number of clients containing 300 samples respectively, and directly investigate how much time it costs regarding the number of clients.\footnote{The evaluation time for all three methods will be doubled for the case of more than 16 clients because the $17^{th}$ client needs to wait until a single-core finished their previous local training. In other words, multi-processing will not be executed when the CPU cores are full.} Table \ref{evaluation_time} shows the evaluation time(in seconds) of one single FL simulation under $\{4,8,12,16,32,64\}$ clients setting. Despite considering the number of simulations($S$) for accurate evaluation, the evaluation time of FedCCEA slowly increases while TMC-SV and LOO need to consume a nontrivial amount of time.

\begin{table}[ht]
  \begin{center}
  \begin{tabular}{c | cccccc}
    \toprule
    \multirow{2}{*}{Evaluation Method} & \multicolumn{6}{c}{Client Number}\\
    \cmidrule(r){2-7}
    & 4 & 8 & 12 & 16 & 32 & 64\\
    \midrule
    FedCCEA  & 8.78s & 10.50s & 12.60s & 14.35s & 22.50s & 38.58s \tabularnewline
    
    TMC-SV & 40.37s & 72.14s & 107.21s & 140.53s & 279.48s & 568.93s \tabularnewline
    
    LOO  & 41.16s & 72.75s & 107.85s & 141.04s & 284.50s &  584.37s  \tabularnewline
    \bottomrule
  \end{tabular}
  \caption{\textbf{Evaluation Time for one single FL simulation.} Note that FedCCEA requires more simulations($S$) to evaluate client contribution accurately.($\times S$ to FedCCEA)} 
  \label{evaluation_time}
  \end{center}
\end{table}

\subsection{Client Contribution Skewness}

\begin{figure}[h]
    \centering
    \subfigure[MNIST]
    {\centering
    \includegraphics[width= 0.9\textwidth, height = 0.09\textheight]{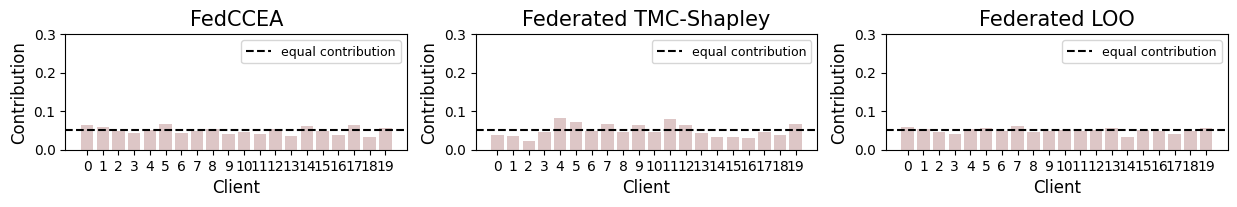}
    \label{contribution_mnist_iid}}

    \subfigure[EMNIST]
    {\centering
    \includegraphics[width= 0.9\textwidth, height = 0.09\textheight]{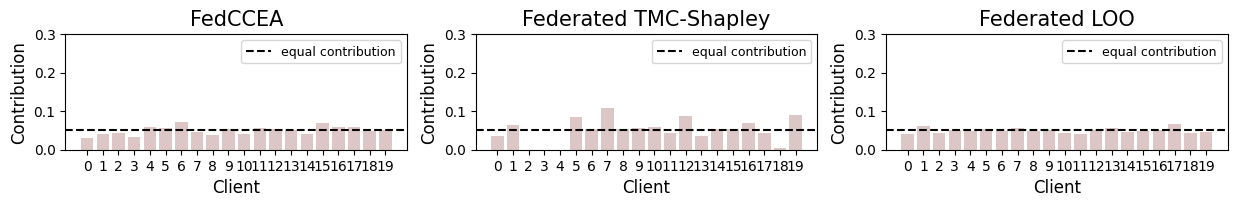}
    \label{contribution_emnist_iid}}
    \caption{\textbf{Client Contribution Index in the IID setting} X-axis: client index, Y-axis: CCI for each client. FedCCEA(left), TMC-SV(center), and LOO(right).}   
    \label{contribution_iid}
\end{figure}

\begin{figure}[t]
    \centering
    \subfigure[MNIST]
    {\centering
    \includegraphics[width= 0.9\textwidth, height = 0.09\textheight]{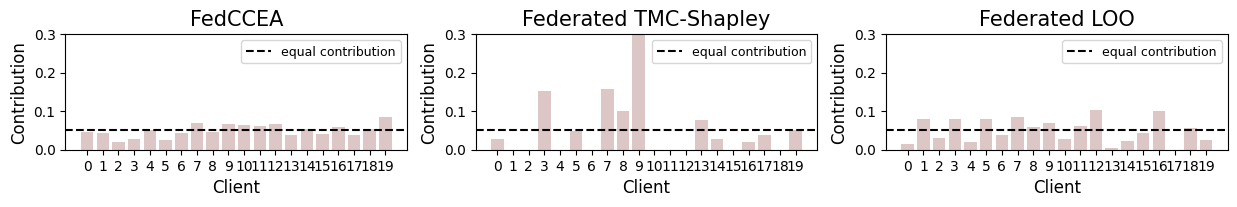}
    \label{contribution_mnist_weak}}

    \subfigure[EMNIST]
    {\centering
    \includegraphics[width= 0.9\textwidth, height = 0.09\textheight]{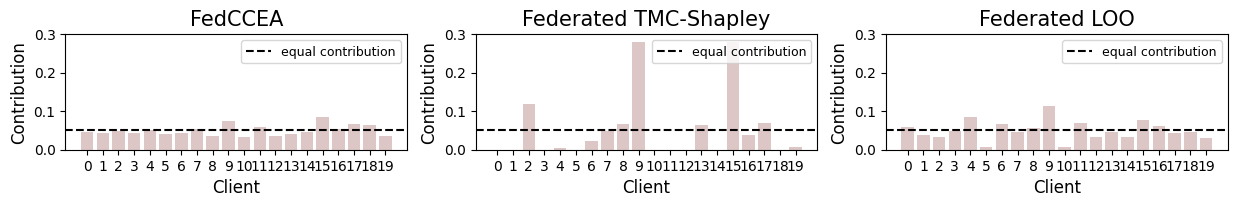}
    \label{contribution_emnist_weak}}
    \caption{\textbf{Client Contribution Index in the weak non-IID setting} X-axis: client index, Y-axis: CCI for each client. FedCCEA(left), TMC-SV(center), and LOO(right).}
    \label{contribution_weak}
\end{figure}

Figure \ref{contribution_iid} and \ref{contribution_weak} illustrate the remaining results of Client Contribution Skewness that are not mentioned in Section \ref{Client Contribution Skewness}. Table \ref{zero_exclusion_all} records the test accuracy after the exclusion of zero contributors measured by three different evaluation methods.

\clearpage

\begin{table}[h]
  \centering
  \begin{tabular}{p{0.15\textwidth}l >{\centering}p{0.15\textwidth}l >{\centering}p{0.15\textwidth}l >{\centering}p{0.15\textwidth}l >{\centering}p{0.15\textwidth}l >{\centering}p{0.15\textwidth}l}
    \toprule
    \multicolumn{3}{c}{} &
    \multicolumn{3}{c}{Client Contribution Evaluation Method} \tabularnewline
    \cmidrule(r){4-6}
    Dataset & Distribution & Base & FedCCEA & TMC-SV & LOO \tabularnewline
    \midrule
     MNIST & IID & 91.12\%  & 91.12\% & 91.12\% & 91.12\% \tabularnewline
           & Weak non-IID & 90.37\% & 90.37\% & 84.87\% & \textbf{90.41\%} \tabularnewline
           & Strong non-IID & 85.60\%  & \textbf{85.60\%} & 74.29\% & 85.29\% \tabularnewline
     \hline
     EMNIST & IID & 77.47\%  & \textbf{77.47\%} & 77.39\% & \textbf{77.47\%} \tabularnewline
           & Weak non-IID & 75.31\%  & \textbf{75.31\%} & 73.01\% & \textbf{75.31\%} \tabularnewline
           & Strong non-IID & 59.64\%  & \textbf{59.64\%} & 50.52\% & 54.65\% \tabularnewline
    \bottomrule
  \end{tabular}
  \caption{\textbf{Exclusion of zero contributors}: Same as Section \ref{Client Contribution Skewness}, implementing federated learning after excluding clients with zero CCIs. The highest test accuracy among three evaluation methods are marked in textbf{bold}.}
  \label{zero_exclusion_all}
\end{table}

\subsection{Client Removal \& Robustness Test to Partial Participation}
For the rest of the Appendix, we show the total results of remaining data distribution settings and noise injection settings that are not reported in the main paper. With same experiments constructed in Section \ref{Client Removal} and \ref{robustness}, the straight line of FedCCEA should retain high accuracy as low-contributed clients are removed while the dashed line of FedCCEA should critically drop the accuracy as high-contributed clients are removed.

\begin{figure}[p]
    \centering
    \subfigure[No Noise]
    {\centering
    \includegraphics[width= 0.27\textwidth, height = 0.18\textheight]{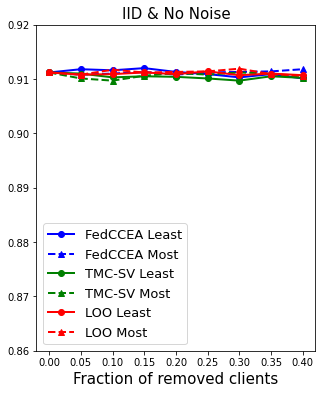}
    \label{cr_mnist_iid_nonoise}}
    \qquad
    \subfigure[40\% Label Noise]
    {\centering
    \includegraphics[width= 0.27\textwidth, height = 0.18\textheight]{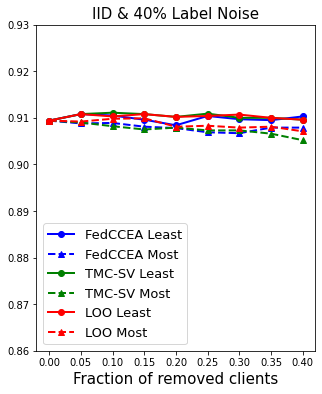}
    \label{cr_mnist_iid_40label}}
    \qquad
    \subfigure[20\% Label Noise]
    {\centering
    \includegraphics[width= 0.27\textwidth, height = 0.18\textheight]{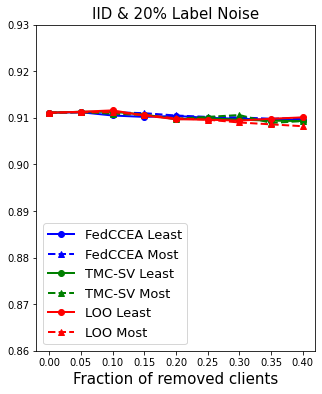}
    \label{cr_mnist_iid_20label}}
    
    \subfigure[10\% Label Noise]
    {\centering
    \includegraphics[width= 0.27\textwidth, height = 0.18\textheight]{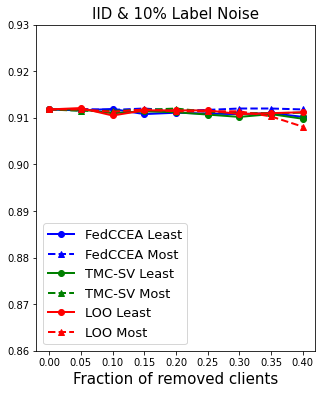}
    \label{cr_mnist_iid_10label}}
    \subfigure[20\% Sample Noise]
    {\centering
    \includegraphics[width= 0.27\textwidth, height = 0.18\textheight]{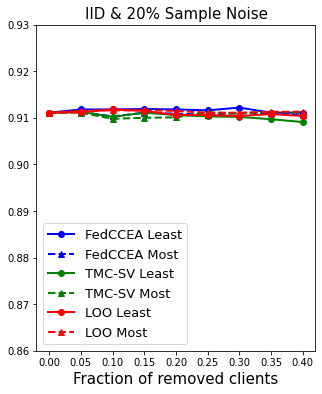}
    \label{cr_mnist_iid_20sample}}

     \caption{\textbf{Client Removal Experiment for MNIST in IID setting}}
    \label{client_removal_mnist_iid}
\end{figure}

\begin{figure}[p]
    \centering
    \subfigure[No Noise]
    {\centering
    \includegraphics[width= 0.27\textwidth, height = 0.18\textheight]{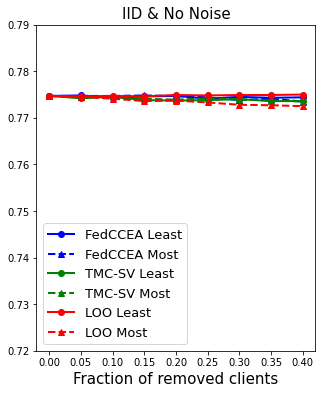}
    \label{cr_emnist_iid_nonoise}}
    \qquad
    \subfigure[40\% Label Noise]
    {\centering
    \includegraphics[width= 0.27\textwidth, height = 0.18\textheight]{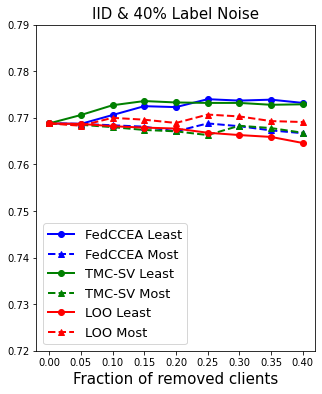}
    \label{cr_emnist_iid_40label}}
    \qquad
    \subfigure[20\% Label Noise]
    {\centering
    \includegraphics[width= 0.27\textwidth, height = 0.18\textheight]{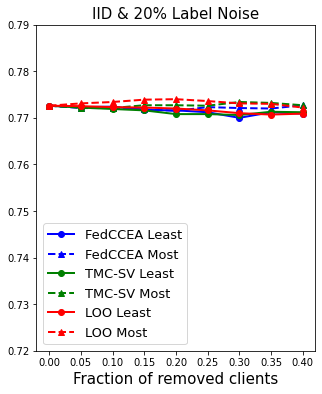}
    \label{cr_emnist_iid_20label}}
    
    \subfigure[10\% Label Noise]
    {\centering
    \includegraphics[width= 0.27\textwidth, height = 0.18\textheight]{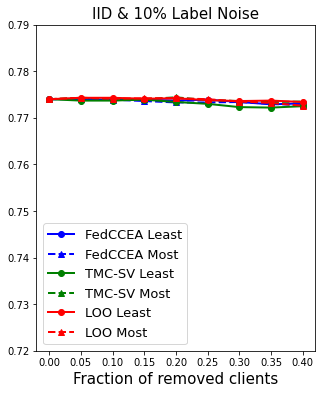}
    \label{cr_emnist_iid_10label}}
    \subfigure[20\% Sample Noise]
    {\centering
    \includegraphics[width= 0.27\textwidth, height = 0.19\textheight]{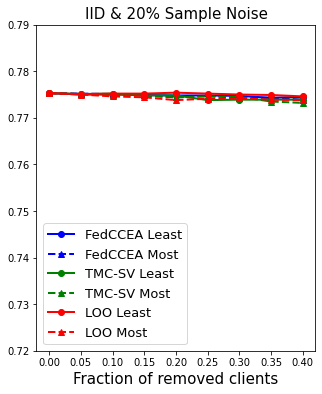}
    \label{cr_emnist_iid_20sample}}
     \caption{\textbf{Client Removal Experiment for EMNIST in IID settings}}
    \label{client_removal_emnist_iid}
\end{figure}

\begin{figure}[p]
    \centering
    \subfigure[No Noise]
    {\centering
    \includegraphics[width= 0.27\textwidth, height = 0.19\textheight]{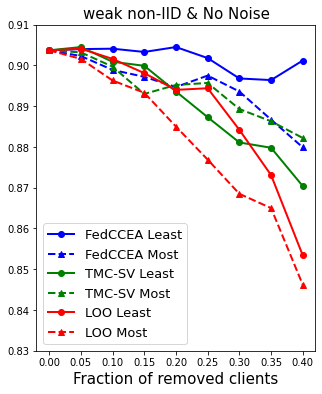}
    \label{cr_mnist_weak_nonoise}}
    \qquad
    \subfigure[40\% Label Noise]
    {\centering
    \includegraphics[width= 0.27\textwidth, height = 0.19\textheight]{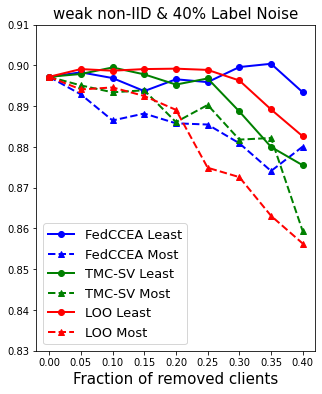}
    \label{cr_mnist_weak_40label}}
    \qquad
    \subfigure[20\% Label Noise]
    {\centering
    \includegraphics[width= 0.27\textwidth, height = 0.19\textheight]{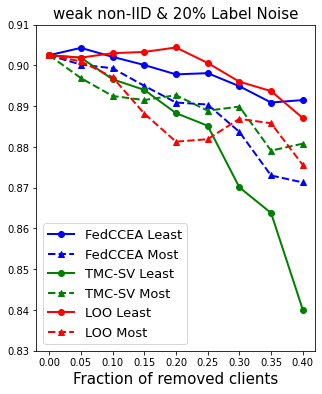}
    \label{cr_mnist_weak_20label}}
    
    \subfigure[10\% Label Noise]
    {\centering
    \includegraphics[width= 0.27\textwidth, height = 0.19\textheight]{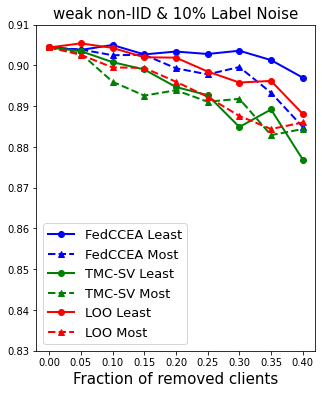}
    \label{cr_mnist_weak_10label}}
    \subfigure[20\% Sample Noise]
    {\centering
    \includegraphics[width= 0.27\textwidth, height = 0.19\textheight]{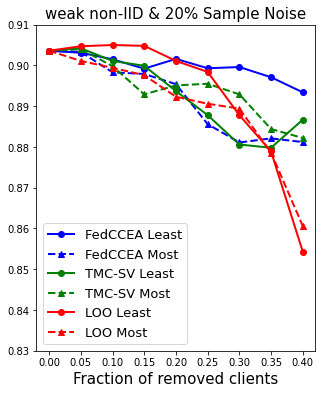}
    \label{cr_mnist_weak_20sample}}

     \caption{\textbf{Client Removal Experiment for MNIST in weak non-IID settings}}
    \label{client_removal_mnist_weak}
\end{figure}

\begin{figure}[p]
    \centering
    \subfigure[No Noise]
    {\centering
    \includegraphics[width= 0.27\textwidth, height = 0.19\textheight]{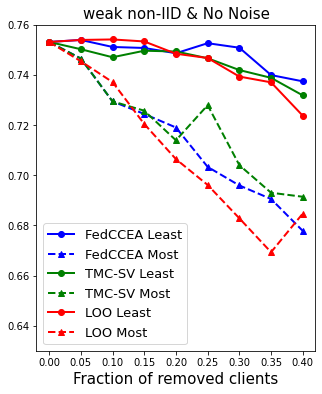}
    \label{cr_emnist_weak_nonoise}}
    \qquad
    \subfigure[40\% Label Noise]
    {\centering
    \includegraphics[width= 0.27\textwidth, height = 0.19\textheight]{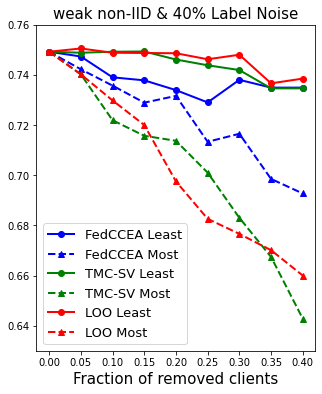}
    \label{cr_emnist_weak_40label}}
    \qquad
    \subfigure[20\% Label Noise]
    {\centering
    \includegraphics[width= 0.27\textwidth, height = 0.19\textheight]{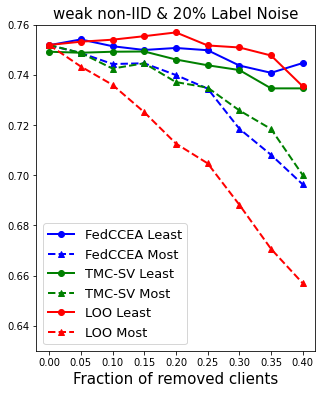}
    \label{cr_emnist_weak_20label}}
    
    \subfigure[10\% Label Noise]
    {\centering
    \includegraphics[width= 0.27\textwidth, height = 0.19\textheight]{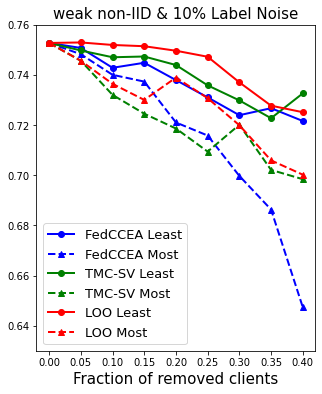}
    \label{cr_emnist_weak_10label}}
    \subfigure[20\% Sample Noise]
    {\centering
    \includegraphics[width= 0.27\textwidth, height = 0.19\textheight]{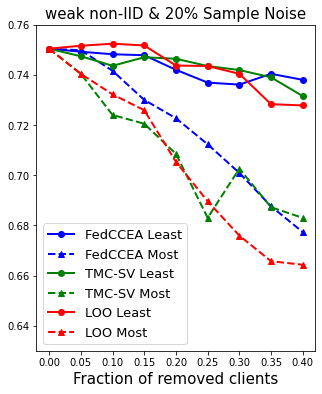}
    \label{cr_emnist_weak_20sample}}
     \caption{\textbf{Client Removal Experiment for EMNIST in weak non-IID settings}}
    \label{client_removal_emnist_weak}
\end{figure}

\begin{figure}[p]
    \centering
    \subfigure[No Noise]
    {\centering
    \includegraphics[width= 0.27\textwidth, height = 0.19\textheight]{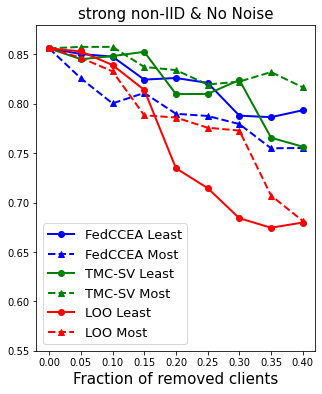}
    \label{cr_mnist_strong_nonoise}}
    \qquad
    \subfigure[40\% Label Noise]
    {\centering
    \includegraphics[width= 0.27\textwidth, height = 0.19\textheight]{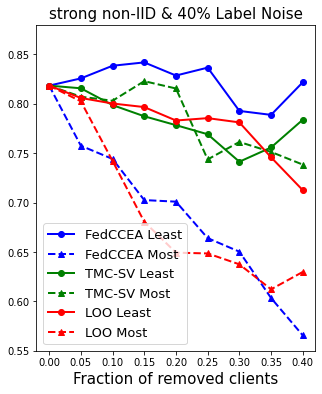}
    \label{cr_mnist_strong_40label}}
    \qquad
    \subfigure[20\% Label Noise]
    {\centering
    \includegraphics[width= 0.27\textwidth, height = 0.19\textheight]{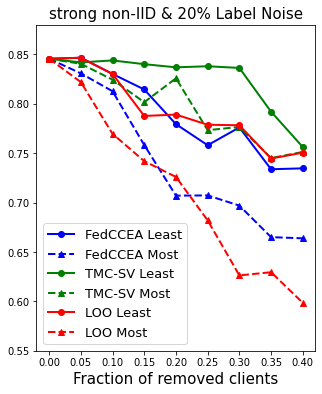}
    \label{cr_mnist_strong_20label}}
    
    \subfigure[10\% Label Noise]
    {\centering
    \includegraphics[width= 0.27\textwidth, height = 0.19\textheight]{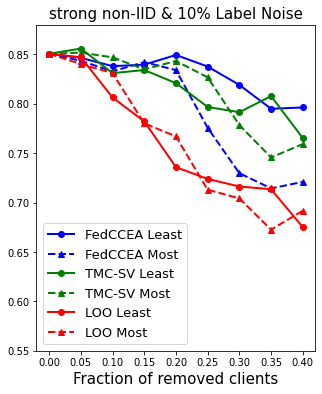}
    \label{cr_mnist_strong_10label}}
    \subfigure[20\% Sample Noise]
    {\centering
    \includegraphics[width= 0.27\textwidth, height = 0.19\textheight]{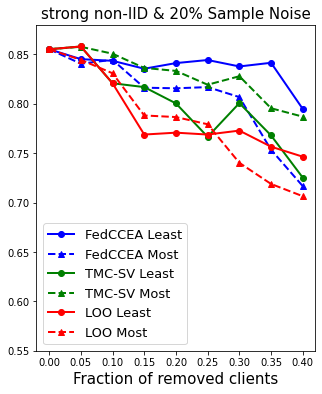}
    \label{cr_mnist_strong_20sample}}

     \caption{\textbf{Client Removal Experiment for MNIST in strong non-IID settings}}
    \label{client_removal_mnist_strong}
\end{figure}

\begin{figure}[p]
    \centering
    \subfigure[No Noise]
    {\centering
    \includegraphics[width= 0.27\textwidth, height = 0.19\textheight]{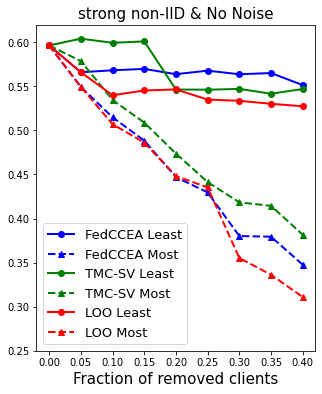}
    \label{cr_emnist_strong_nonoise}}
    \qquad
    \subfigure[40\% Label Noise]
    {\centering
    \includegraphics[width= 0.27\textwidth, height = 0.19\textheight]{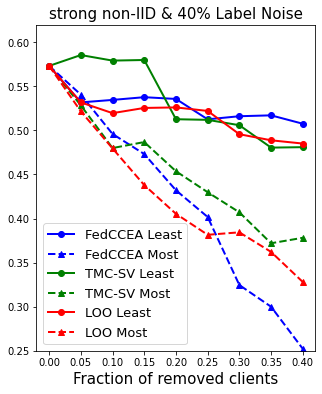}
    \label{cr_emnist_strong_40label}}
    \qquad
    \subfigure[20\% Label Noise]
    {\centering
    \includegraphics[width= 0.27\textwidth, height = 0.19\textheight]{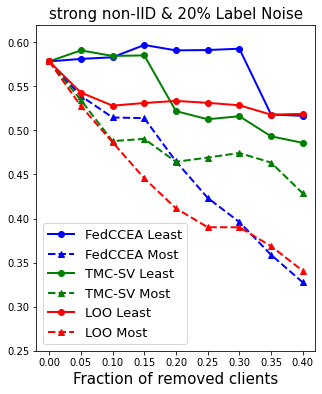}
    \label{cr_emnist_strong_20label}}
    
    \subfigure[10\% Label Noise]
    {\centering
    \includegraphics[width= 0.27\textwidth, height = 0.19\textheight]{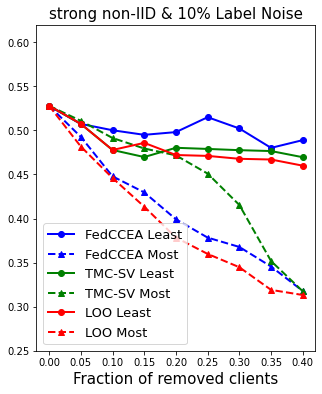}
    \label{cr_emnist_strong_10label}}
    \subfigure[20\% Sample Noise]
    {\centering
    \includegraphics[width= 0.27\textwidth, height = 0.19\textheight]{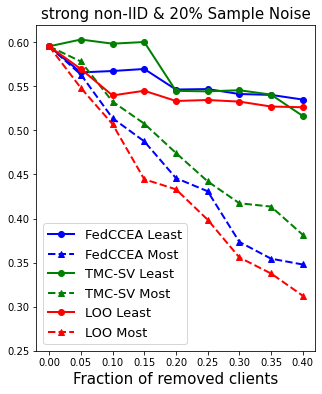}
    \label{cr_emnist_strong_20sample}}
     \caption{\textbf{Client Removal Experiment for EMNIST in strong non-IID settings}}
    \label{client_removal_emnist_strong}
\end{figure}

\begin{figure}[p]
    \centering
    \subfigure[No Noise]
    {\centering
    \includegraphics[width= 0.27\textwidth, height = 0.19\textheight]{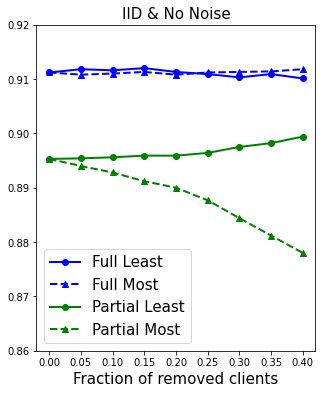}
    \label{rob_mnist_iid_nonoise}}
    \qquad
    \subfigure[40\% Label Noise]
    {\centering
    \includegraphics[width= 0.27\textwidth, height = 0.19\textheight]{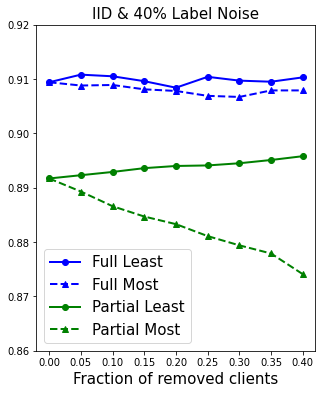}
    \label{rob_mnist_iid_40label}}
    \qquad
    \subfigure[20\% Label Noise]
    {\centering
    \includegraphics[width= 0.27\textwidth, height = 0.19\textheight]{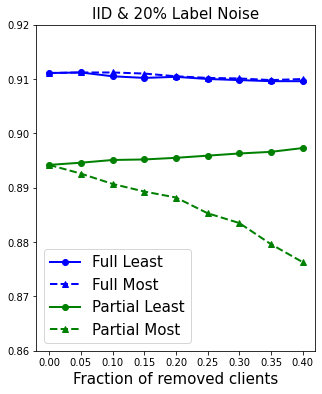}
    \label{rob_mnist_iid_20label}}
    
    \subfigure[10\% Label Noise]
    {\centering
    \includegraphics[width= 0.27\textwidth, height = 0.19\textheight]{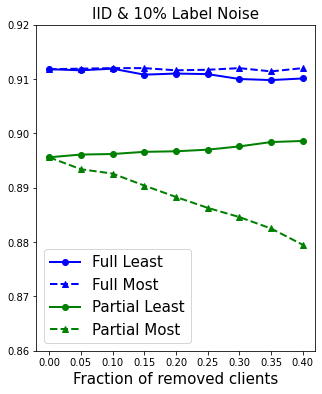}
    \label{rob_mnist_iid_10label}}
    \subfigure[20\% Sample Noise]
    {\centering
    \includegraphics[width= 0.27\textwidth, height = 0.19\textheight]{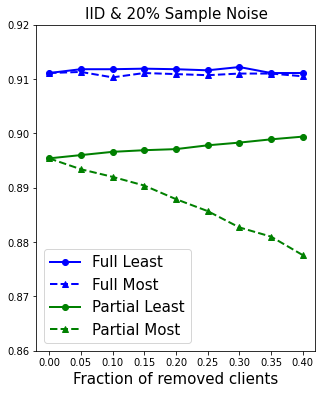}
    \label{rob_mnist_iid_20sample}}

     \caption{\textbf{Robustness Test to Partial Participation for MNIST in IID settings}}
    \label{robustness_mnist_iid}
\end{figure}

\begin{figure}[p]
    \centering
    \subfigure[No Noise]
    {\centering
    \includegraphics[width= 0.27\textwidth, height = 0.19\textheight]{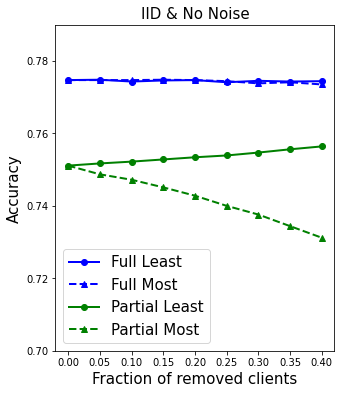}
    \label{rob_emnist_iid_nonoise}}
    \qquad
    \subfigure[40\% Label Noise]
    {\centering
    \includegraphics[width= 0.27\textwidth, height = 0.19\textheight]{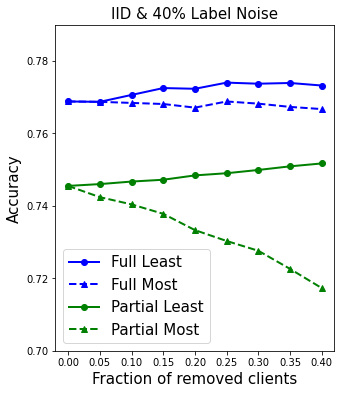}
    \label{rob_emnist_iid_40label}}
    \qquad
    \subfigure[20\% Label Noise]
    {\centering
    \includegraphics[width= 0.27\textwidth, height = 0.19\textheight]{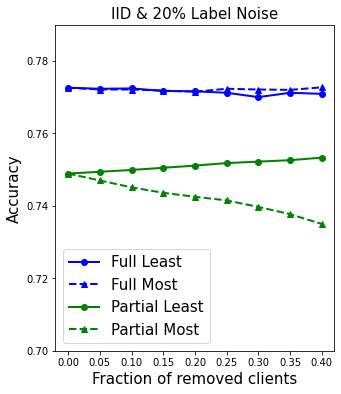}
    \label{rob_emnist_iid_20label}}
    
    \subfigure[10\% Label Noise]
    {\centering
    \includegraphics[width= 0.27\textwidth, height = 0.19\textheight]{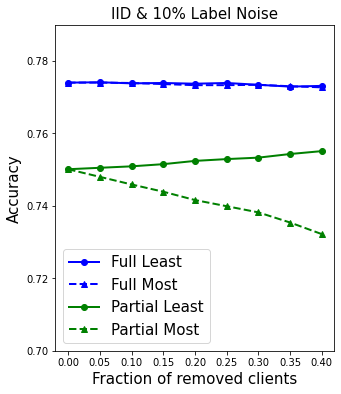}
    \label{rob_emnist_iid_10label}}
    \subfigure[20\% Sample Noise]
    {\centering
    \includegraphics[width= 0.27\textwidth, height = 0.19\textheight]{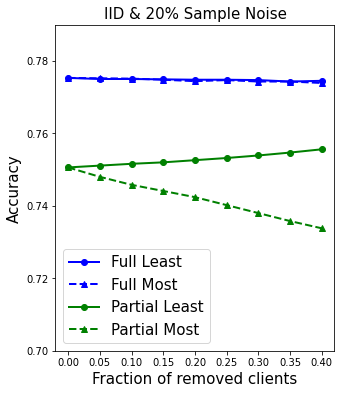}
    \label{rob_emnist_iid_20sample}}

     \caption{\textbf{Robustness Test to Partial Participation for EMNIST in IID settings}}
    \label{robustness_emnist_iid}
\end{figure}

\begin{figure}[p]
    \centering
    \subfigure[No Noise]
    {\centering
    \includegraphics[width= 0.27\textwidth, height = 0.19\textheight]{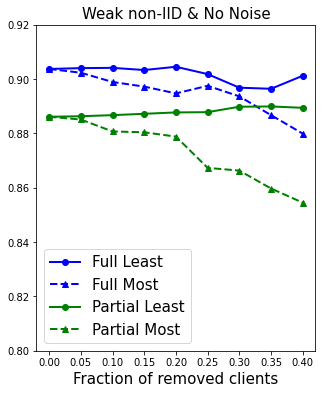}
    \label{rob_mnist_weak_nonoise}}
    \qquad
    \subfigure[40\% Label Noise]
    {\centering
    \includegraphics[width= 0.27\textwidth, height = 0.19\textheight]{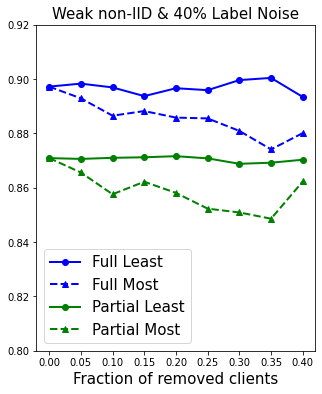}
    \label{rob_mnist_weak_40label}}
    \qquad
    \subfigure[20\% Label Noise]
    {\centering
    \includegraphics[width= 0.27\textwidth, height = 0.19\textheight]{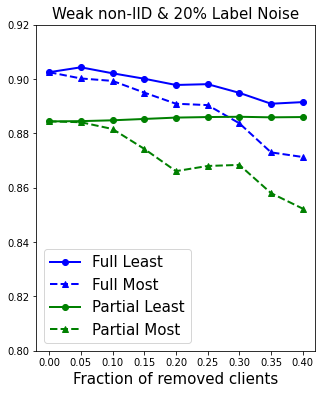}
    \label{rob_mnist_weak_20label}}
    
    \subfigure[10\% Label Noise]
    {\centering
    \includegraphics[width= 0.27\textwidth, height = 0.19\textheight]{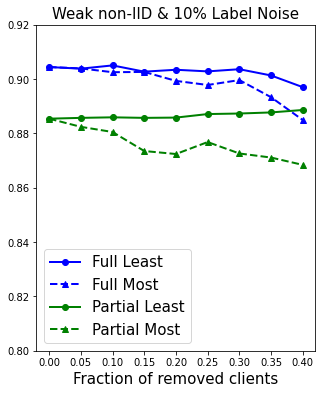}
    \label{rob_mnist_weak_10label}}
    \subfigure[20\% Sample Noise]
    {\centering
    \includegraphics[width= 0.27\textwidth, height = 0.19\textheight]{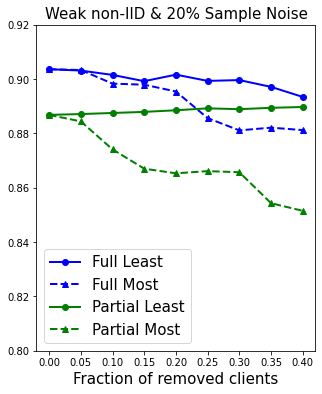}
    \label{rob_mnist_weak_20sample}}

     \caption{\textbf{Robustness Test to Partial Participation for MNIST in weak non-IID settings}}
    \label{robustness_mnist_weak}
\end{figure}

\begin{figure}[p]
    \centering
    \subfigure[No Noise]
    {\centering
    \includegraphics[width= 0.27\textwidth, height = 0.19\textheight]{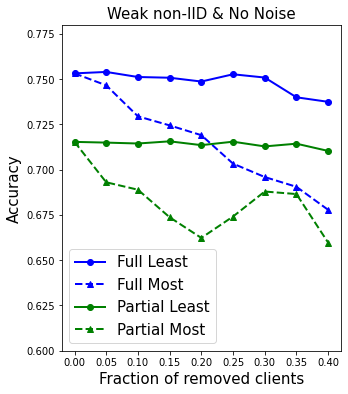}
    \label{rob_emnist_weak_nonoise}}
    \qquad
    \subfigure[40\% Label Noise]
    {\centering
    \includegraphics[width= 0.27\textwidth, height = 0.19\textheight]{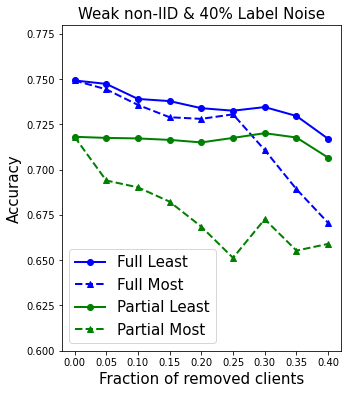}
    \label{rob_emnist_weak_40label}}
    \qquad
    \subfigure[20\% Label Noise]
    {\centering
    \includegraphics[width= 0.27\textwidth, height = 0.19\textheight]{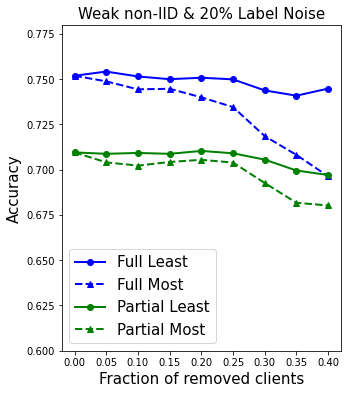}
    \label{rob_emnist_weak_20label}}
    
    \subfigure[10\% Label Noise]
    {\centering
    \includegraphics[width= 0.27\textwidth, height = 0.19\textheight]{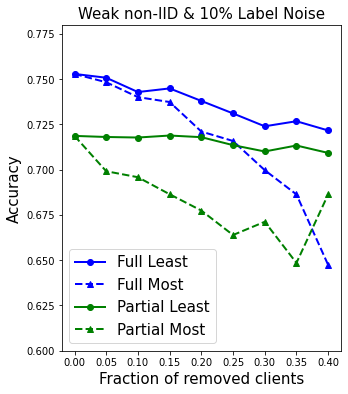}
    \label{rob_emnist_weak_10label}}
    \subfigure[20\% Sample Noise]
    {\centering
    \includegraphics[width= 0.27\textwidth, height = 0.19\textheight]{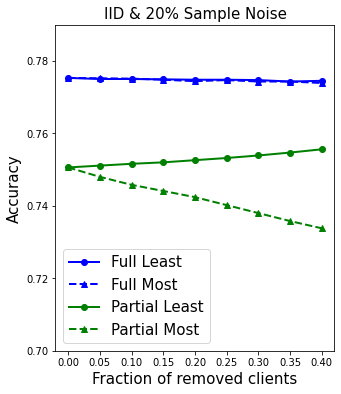}
    \label{rob_emnist_weak_20sample}}

     \caption{\textbf{Robustness Test to Partial Participation for EMNIST in weak non-IID settings}}
    \label{robustness_emnist_weak}
\end{figure}

\begin{figure}[p]
    \centering
    \subfigure[No Noise]
    {\centering
    \includegraphics[width= 0.27\textwidth, height = 0.19\textheight]{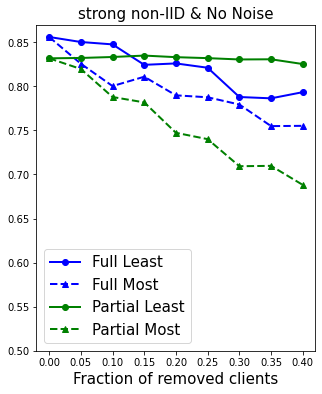}
    \label{rob_mnist_strong_nonoise}}
    \qquad
    \subfigure[40\% Label Noise]
    {\centering
    \includegraphics[width= 0.27\textwidth, height = 0.19\textheight]{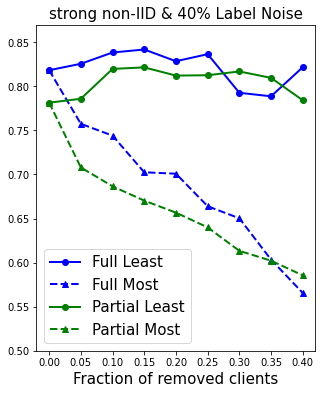}
    \label{rob_mnist_strong_40label}}
    \qquad
    \subfigure[20\% Label Noise]
    {\centering
    \includegraphics[width= 0.27\textwidth, height = 0.19\textheight]{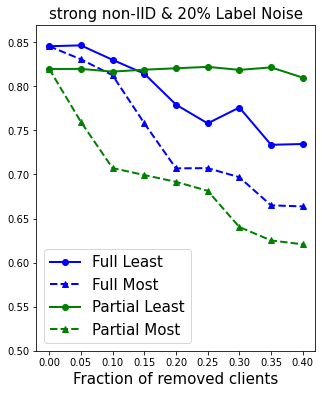}
    \label{rob_mnist_strong_20label}}
    
    \subfigure[10\% Label Noise]
    {\centering
    \includegraphics[width= 0.27\textwidth, height = 0.19\textheight]{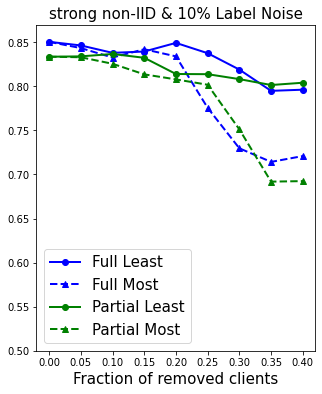}
    \label{rob_mnist_strong_10label}}
    \subfigure[20\% Sample Noise]
    {\centering
    \includegraphics[width= 0.27\textwidth, height = 0.19\textheight]{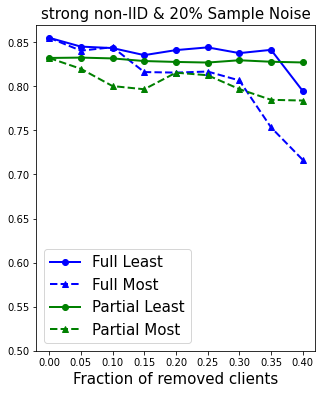}
    \label{rob_mnist_strong_20sample}}

     \caption{\textbf{Robustness Test to Partial Participation for MNIST in strong non-IID settings}}
    \label{robustness_mnist_strong}
\end{figure}

\begin{figure}[p]
    \centering
    \subfigure[No Noise]
    {\centering
    \includegraphics[width= 0.27\textwidth, height = 0.19\textheight]{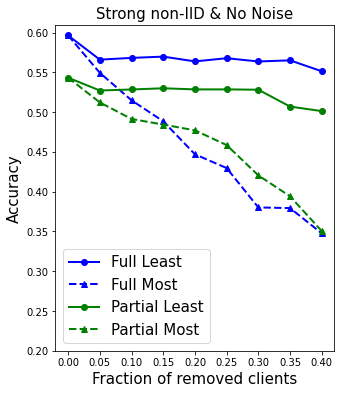}
    \label{rob_emnist_strong_nonoise}}
    \qquad
    \subfigure[40\% Label Noise]
    {\centering
    \includegraphics[width= 0.27\textwidth, height = 0.19\textheight]{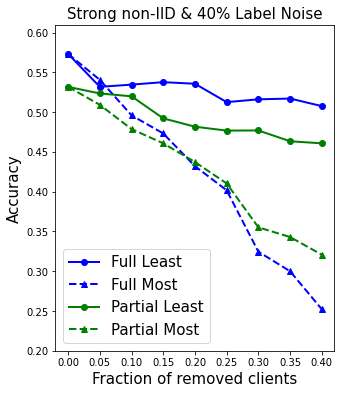}
    \label{rob_emnist_strong_40label}}
    \qquad
    \subfigure[20\% Label Noise]
    {\centering
    \includegraphics[width= 0.27\textwidth, height = 0.19\textheight]{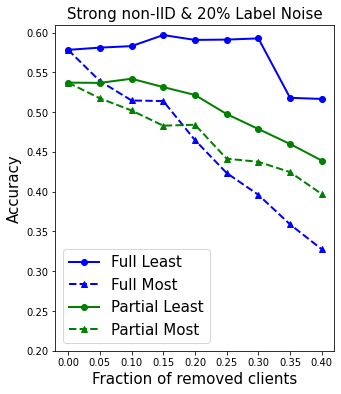}
    \label{rob_emnist_strong_20label}}
    
    \subfigure[10\% Label Noise]
    {\centering
    \includegraphics[width= 0.27\textwidth, height = 0.19\textheight]{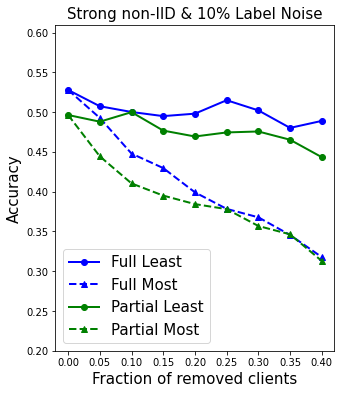}
    \label{rob_emnist_strong_10label}}
    \subfigure[20\% Sample Noise]
    {\centering
    \includegraphics[width= 0.27\textwidth, height = 0.19\textheight]{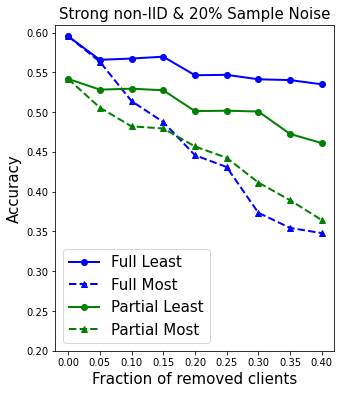}
    \label{rob_emnist_strong_20sample}}

     \caption{\textbf{Robustness Test to Partial Participation for EMNIST in strong non-IID settings}}
    \label{robustness_emnist_strong}
\end{figure}
\end{appendices}
\end{document}